\documentclass[10pt,twocolumn,letterpaper]{article}

\usepackage[pagenumbers]{arxiv} 

\usepackage{graphicx}
\usepackage{amsmath}
\usepackage{amssymb}
\usepackage{booktabs}
\usepackage{multirow}
\usepackage{enumitem}
\usepackage{array}

\usepackage[pagebackref,breaklinks,colorlinks]{hyperref}

\usepackage[capitalize]{cleveref}
\crefname{section}{Sec.}{Secs.}
\Crefname{section}{Section}{Sections}
\Crefname{table}{Table}{Tables}
\crefname{table}{Tab.}{Tabs.}

\begin{document}

\title{SOWA: Adapting Hierarchical Frozen Window Self-Attention to Visual-Language Models for Better Anomaly Detection}

\author{Zongxiang Hu\footnotemark[1] \footnotemark[2]\\
\and
Zhaosheng Zhang\footnotemark[1]\\
}
\maketitle

\renewcommand{\thefootnote}{\fnsymbol{footnote}}
\footnotetext[1]{These authors contributed equally to this work.}
\footnotetext[2]{Corresponding author. {\tt\small huzongxiang1991@gmail.com}}
\renewcommand{\thefootnote}{\arabic{footnote}}

\begin{abstract}
Visual anomaly detection is essential in industrial manufacturing, yet traditional methods often rely heavily on extensive normal datasets and task-specific models, limiting their scalability. Recent advancements in large-scale vision-language models have significantly enhanced zero- and few-shot anomaly detection. However, these approaches may not fully leverage hierarchical features, potentially overlooking nuanced details crucial for accurate detection. To address this, we introduce a novel window self-attention mechanism based on the CLIP model, augmented with learnable prompts to process multi-level features within a Soldier-Officer Window Self-Attention (SOWA) framework. Our method has been rigorously evaluated on five benchmark datasets, achieving superior performance by leading in 18 out of 20 metrics, setting a new standard against existing state-of-the-art techniques.
\end{abstract}

\section{Introduction}
\label{sec:intro}

\begin{figure}[t]
\centering
\includegraphics[width=0.9\columnwidth]{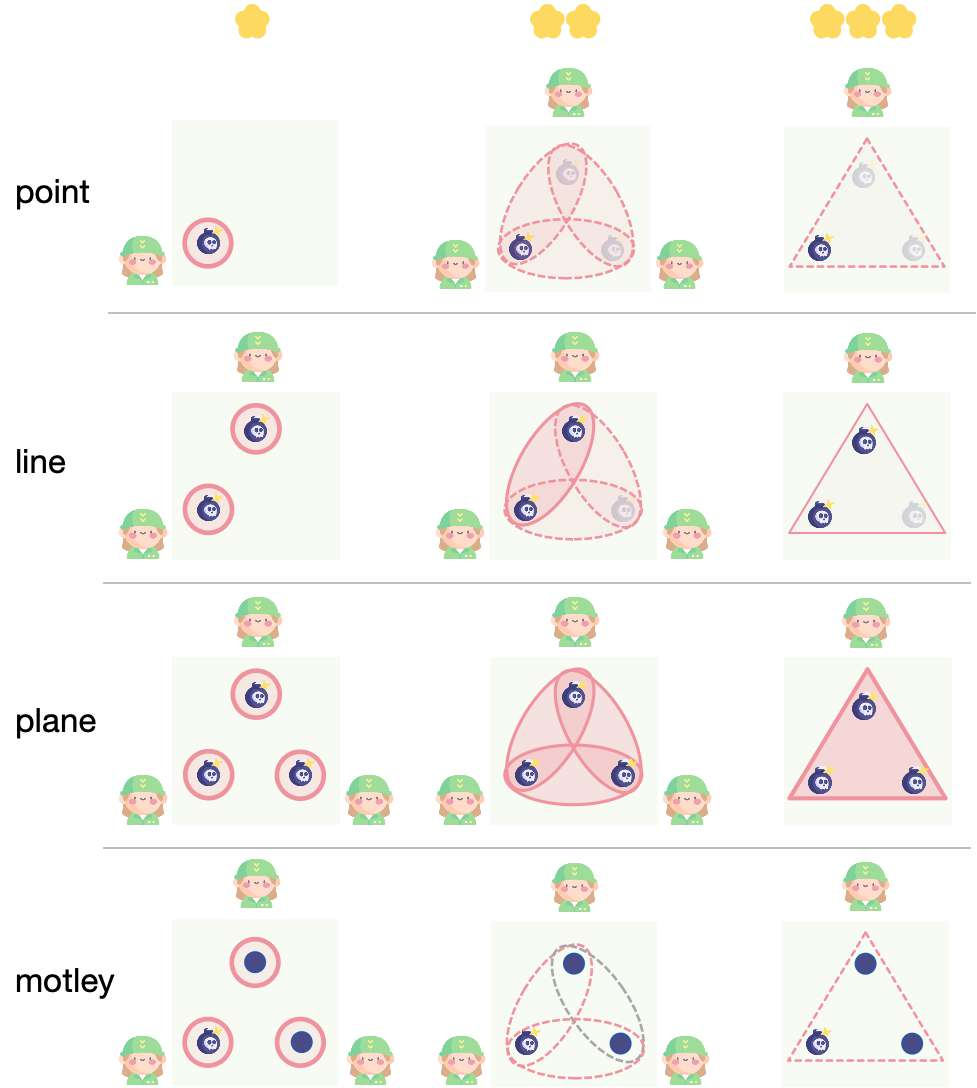}
\caption{A diagram illustrating the conceptual handling of different abnormal image feature patterns (point, line, plane, motley) with the number of stars indicating the depth of the ViT.
\textbf{Point:} Few local textures and anomalies, best detected in shallow layers, as deeper layers diffuse these local features, making fine-grained anomalies harder to detect.
\textbf{Line:} Involves both local features and large-scale observation, best processed in middle layers.
\textbf{Plane:} Handles large areas of features, processed in deep layers where global properties and shapes are captured.
\textbf{Motley:} Rich in local textures and anomalies, requiring a comprehensive approach from shallow to deep layers.}
\label{fig:figure 1}
\end{figure}

Visual anomaly classification (AC) and segmentation (AS) are crucial tasks in industrial manufacturing, aimed at identifying and localizing defects within images \cite{bae_pni_2023, bergmann_beyond_2022, roth_towards_2022, liu_simplenet_2023, batzner_efficientad_2024, wyatt_anoddpm_2022}. The primary challenge of these tasks is the rarity and variability of defects, leading to a scarcity of representative anomaly samples in training data. Most existing methods emphasize one-class or unsupervised anomaly detection, requiring a large volume of normal images and task-specific models, which limits scalability across diverse tasks \cite{roth_towards_2022, deng_anomaly_2022, bergmann_beyond_2022, cohen_sub-image_2021, defard_padim_2021, li_cutpaste_2021,ristea_self-supervised_2022, zavrtanik_draem_2021, zou_spot--difference_2022}. Although high accuracy can be achieved with ample normal data \cite{roth_towards_2022, bergmann_mvtec_2019, defard_padim_2021}, performance declines significantly with fewer normal samples \cite{huang_registration_2022, sheynin_hierarchical_2021, zou_spot--difference_2022}, making data scarcity a key bottleneck for improved performance. To reduce reliance on extensive data, we need to seek more robust and advanced methods to tackle this problem.

Vision-language models (VLMs) \cite{radford_learning_2021, jia_scaling_2021, gu_open-vocabulary_2021}, such as CLIP, have demonstrated strong performance in zero- and few-shot classification tasks through  large-scale training on vision-language annotated pairs via contrastive learning  \cite{chen_simple_2020}. This training enables VLMs to capture broad concepts and generalize to new tasks. For instance, WinCLIP \cite{jeong_winclip_2023}, an extension of CLIP, applies a multi-scale window-based strategy for segmentation, using text prompts to guide anomaly detection and effectively performing zero- and few-shot anomaly detection without extensive category-specific training. Similarly, April-GAN \cite{chen_april-gan_2023} introduces additional linear layers to map image features obtained from CLIP's image encoder into the text feature space, facilitating precise similarity comparisons and generating detailed anomaly maps. The main advantage of CLIP-based approaches lie in their strong generalization capabilities, achieved through large-scale vision-language pretraining. By leveraging language to provide contextual understanding, these models can effectively differentiate between normal and anomalous states, even with limited normal samples, which is particularly valuable when anomaly samples are difficult to obtain.

However, these approaches also have limitations. First, they often underutilize the hierarchical features captured by the CLIP vision transformer (ViT) \cite{dosovitskiy_image_2020}, potentially missing nuanced details crucial for accurate anomaly detection. Second, although effective in using text prompts, the fixed encoded prompts are primarily designed for final global features, which may not optimally integrate local features from earlier layers. These limitations can create challenges in detecting anomalies that vary significantly in scale and context.

To address these limitations and fully leverage multi-level features, we classify anomalous features into four patterns based on their size and depth: point, line, plane, and motley, as illustrated in Figure \ref{fig:figure 1}. Traditional self-attention mechanisms \cite{vaswani_attention_2017}, relying on global self-attention \cite{xing_less_2024}, struggle with these complex patterns and their combinations. Thus, we conceptualize the operational patterns of the model as a \textbf{Soldier-Officer} model: different levels result in varied fields of view. Shallow layers act as soldiers, excelling at identifying local features such as points, while middle layers are better suited for recognizing features with small local depth but large span, like lines. The motley pattern indicates that shallow features, being less influenced by text supervision, have weaker semantic perceptions of anomalies and normalities. Inspired by the convolutional receptive field, we propose window self-attention to capture points and lines. Additionally, fixed encoded prompts are inadequate for multi-level features, so we employ learnable prompts through prompt tuning, akin to soldiers and officers receiving different orders. These innovations lead to our Soldier-Officer Window self-Attention (SOWA) framework, which processes multi-level features of ViT through window-based self-attention. By injecting and freezing CLIP's attention weights into window self-attention, we inherit its feature extraction capabilities while integrating broader context. Introducing learnable prompts overcomes the limitations of manual prompt engineering, enhancing the detection of anomalies across varying scales and contexts. This design enables adaptive focus on relevant features at different levels, leveraging each layer's strengths for improved anomaly detection.

The main contributions are summarized below:
\begin{itemize}[noitemsep, topsep=0pt, leftmargin=*]
\item We propose a carefully crafted framework that leverages the strengths of vision-language models for anomaly detection. 
\item The method is evaluated on multiple benchmark datasets, demonstrating superior performance compared to existing state-of-the-art techniques in both anomaly classification and segmentation tasks. 
\item The model’s operational mechanism is validated, providing valuable insights that can inform future research.  
\end{itemize}

\section{Related Work}
\label{sec:related_work}

\noindent\textbf{Anomaly Classification and Segmentation.} Traditional anomaly detection (AD) methods predominantly focus on one-class classification frameworks \cite{ruff_deep_2018, defard_padim_2021, li_cutpaste_2021, yi_patch_2021, gudovskiy_cflow-ad_2022, zavrtanik_draem_2021}, which rely on extensive datasets of normal images to model normality and detect deviations as anomalies. While effective on benchmarks like MVTec-AD \cite{bergmann_mvtec_2019}, these methods are limited by their dependence on large quantities of normal data.

\noindent\textbf{Few-Shot Anomaly Detection.} Few-shot anomaly detection (FSAD) aims to identify anomalies with minimal normal samples, addressing the practical limitation of data scarcity. Traditional FSAD methods often require re-training or fine-tuning, which hampers their adaptability to new domains. Distance-based methods like SPADE \cite{cohen_sub-image_2021}, PaDiM \cite{defard_padim_2021}, and PatchCore \cite{roth_towards_2022} overcome this by leveraging pre-trained representations to calculate anomaly scores without additional training. WinCLIP has advanced this field by applying VLMs for zero-shot and few-shot AD, utilizing multi-scale windowing and text prompts within CLIP.

\noindent\textbf{Prompt Learning in CLIP.} The success of CLIP in zero-shot classification has led to advancements in prompt learning \cite{jiang_how_2020, rao_denseclip_2022, shin_autoprompt_2020, li_prefix-tuning_2021, zhou_conditional_2022, li_promptad_2024}, where DualCoOp \cite{sun_dualcoop_2022} uses a pair of differentiable prompts to provide positive and negative contexts for the target class. Instead of relying on hand-crafted prompts, the dual prompts naturally result in a positive and a negative classifier, making it easier to determine the presence of the target class in an image by comparing their scores. This method avoids fine-tuning the entire vision-language model and focuses only on learning the prompts, significantly enhancing efficiency when adapting to different datasets \cite{zhou_anomalyclip_2023}.

\begin{figure*}[t]
\centering
\includegraphics[width=\textwidth]{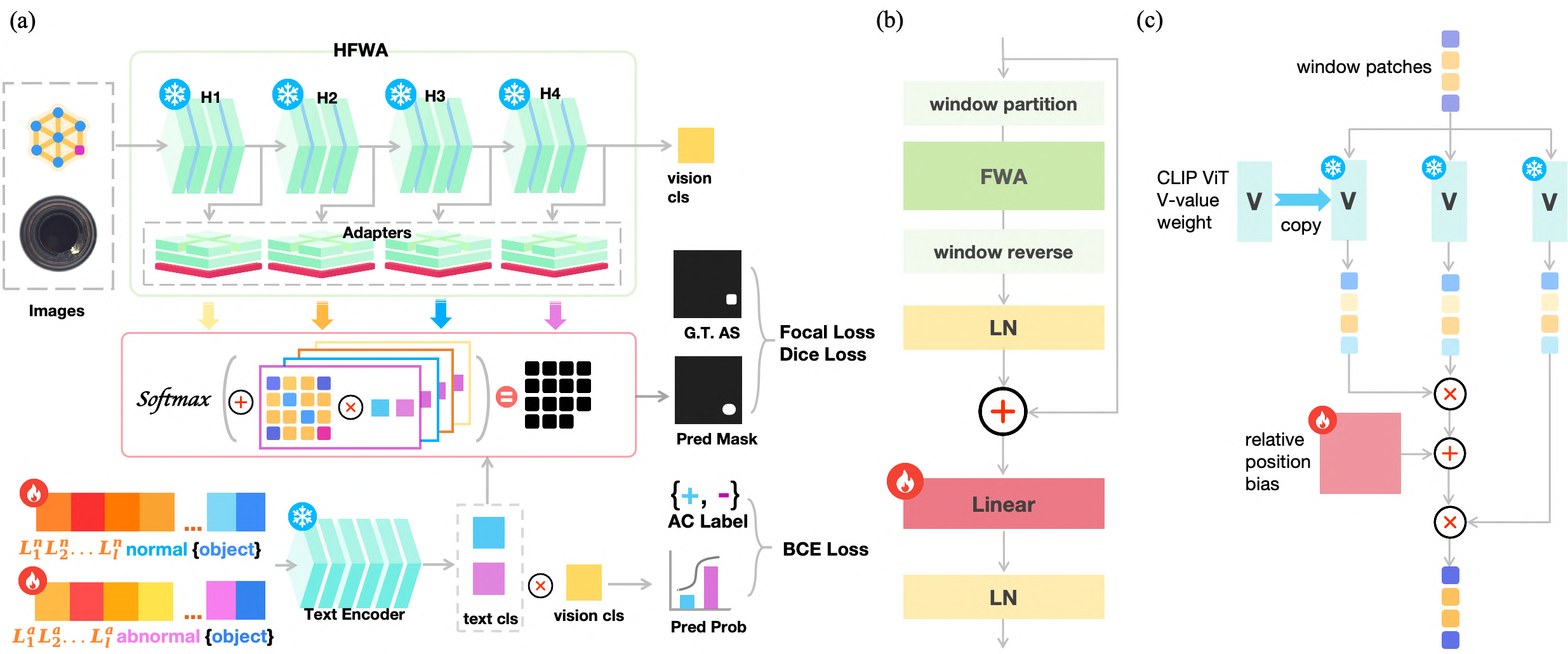}
\caption{(a) The architecture of Hierarchical Frozen Window Self-Attention (HFWA) model. Block H1 to H4 represent the four feature extraction stages of CLIP ViT; (b) Detailed structure of FWA (Frozen Window Attention) adapter; (c) Detailed structure of FWA.}
\label{fig:figure 2}
\end{figure*}

\section{Problem Defination.} 
\label{sec:define}

Let $D_{train}=\{(x_i, c_i, S_i)\}$ denote a dataset containing annotated samples. Each sample is given as a tuple $(x_i, c_i, S_i)$, where $x_i$ is an image of size $h \times w$, $c_i \in \{-1, +1\}$ represents the image-level anomaly classification label, and $S_i \in \{-1, +1\}^{h \times w}$ is a pixel-level anomaly segmentation mask of size $h \times w$, with `1' indicating anomaly, and `-1' indicating normal. Given an image $x$, we aim to predict both image-level ($c$) and pixel-level anomalies ($S$). The image-level prediction is a binary classification that indicates whether anomalies are present, while the pixel-level prediction pinpoints the specific locations of anomalies within the image. In zero-shot learning, the test dataset includes data from domains distinct from the training dataset to assess the model's generalization ability. For few-shot learning, a reference set $T^{\text ref} = \{D_{\alpha}^{\text ref}\}$ contains small reference datasets from the same domain as test samples. For each test sample $x_{\text test}$ in $D_{test}$, $M_{\text adapt}$ uses $K$ randomly selected reference images from the corresponding $D_{\alpha}^{\text ref}$ in $T^{\text ref}$ to evaluate $x_{\text test}$.

\section{SOWA: Soldier-Officer Window Attention}
\label{sec:method}

In the SOWA model, We introduce the Hierarchical Frozen Window Self-Attention (HFWA) architecture to capture and balance hierarchical features, which adapts multi-level output features of various Transformer layers to anomaly patterns.

\subsection{Hierarchical Frozen Window Self-Attention}
\label{sec:hfwa}

The overall structure of the HFWA model is illustrated in Figure \ref{fig:figure 2}a, which is based on the CLIP architecture but incorporates multi-level FWA (Frozen Window Self-Attention) adapters at various ViT layers:  capturing local information in shallow layers, balancing local and global information in middle layers, and integrating global semantic information in deep layers. For an image $x \in \mathbb{R}^{h \times w \times 3}$, the CLIP visual encoder processes it through four sequential stages ($H_1$ to $H_4$),  transforming the image into a CLIP visual representation $F \in \mathbb{R}^{L \times C}$, where $L$ represents the length of visual sequences and $C$ denotes the feature dimension. The output of the four stages ($H_1$ to $H_4$), denoted as $F_{H_i} \in \mathbb{R}^{L \times C}$ , $i \in \{1,2,3,4\}$, encapsulate the hierarchical features obtained at each level.

The FWA adapters, added after the layer of each CLIP stage ($H_1$ to $H_4$), employ a frozen windowed self-attention mechanism. The visual feature adaptation involves four FWA (Frozen Window Self-Attention) adapters $FWA_{i}(\cdot)$. Each FWA adapter $FWA_{i}(\cdot)$ is applied to the CLIP feature $F_{H_i}$, where $i \in \{1, 2, 3, 4\}$, to process features using a windowed self-attention mechanism. The process is defined as:
\begin{equation}
F_{H_i}^* = FWA_i(F_{H_i})
\tag{1}
\end{equation}
After processing through FWA adapters, the features are denoted as $F_{H_i}^*$. The transformed features from all four stages are then fused to form the final feature representation. 

The fusion of these hierarchical features is performed by combining them with the prompt learnable textual features $F_{\text{text}}$, and can be expressed as:
\begin{equation}
F_{\text{fusion}} = \sum_{i=1}^{4} \alpha_{i} F_{H_i}^* F^T_{\text{text}} \tag{2}
\end{equation}
Here, $\alpha_i$ are weights of hierarchical features and set to 1.0 by default. This combined feature map $F_{\text{fusion}}$  captures hierarchical information from all stages of the visual encoder, enhanced with the textual features.

\subsection{Dual Learnable Prompts}
\label{sec:dual_coop}

Commonly used text prompt templates in CLIP, such as "A photo of a [cls]," primarily focus on object semantics and fail to capture anomaly and normal semantics needed for querying corresponding visual embeddings \cite{radford_learning_2021}. Moreover, due to the semantic mapping performed on the shallow features of ViT, the textual features directly encoded by the text encoder are not suitable. To address these, we introduce learnable text prompt templates that incorporate anomaly semantics \cite{sun_dualcoop_2022, li_promptad_2024}. Instead of manually defining templates for specific anomalies, which is impractical due to the unknown and diverse nature of anomalies, we adopt a generic approach using the template "abnormal [cls]." These templates are fine-tuned with auxiliary anomaly detection (AD)-relevant data to generate textual embeddings that effectively discriminate between normality and abnormality. Furthermore, following object-agnostic text prompt learning \cite{zhou_anomalyclip_2023}, we replace specific classes with the fixed term "object”. Finally, a pair of prompts are given to the text encoder as follows::
\begin{equation}
\begin{aligned}
p_n &= [L^n_1][L^n_2] \ldots [L^n_E][\text{normal}][\text{object}] \\
p_a &= [L^a_1][L^a_2] \ldots [L^a_l][\text{abnormal}][\text{object}]
\end{aligned}
\tag{3}
\end{equation}
where $[L_i]  (i \in 1, \ldots, l)$ are learnable word embeddings in normality and abnormality text prompt templates, respectively.

\subsection{FWA Adapter}
\label{sec:adapter}

The structure of the Frozen Window Self-Attention (FWA) adapter is detailed Figure \ref{fig:figure 2}b. The process begins by dividing image features into multiple windows. Self-attention is then applied within each window to enhance local features with richer contextual information. The processed window features are recombined into the overall feature map, creating strong global associations. A trainable linear layer aligns these features with learnable text features and adjusts the weight of FWA features from different levels. The FWA operates as follows:
\begin{equation}
\begin{aligned}
&F_{H_i}^{\text{partitioned}} = \text{WindowPartition}(F_{H_i}) \\ &F_{H_i}^{\text{attended}} = \text{Attention}(F_{H_i}^{\text{partitioned}}) \\ &F_{H_i}^{\text{reversed}} = \text{WindowReverse}(F_{H_i}^{\text{attended}}) \\ &F_{H_i}^{*} = P(F_{H_i}^{\text{reversed}})
\end{aligned}
\tag{4}
\end{equation}

The FWA enhances the CLIP ViT model's capability for anomaly detection by introducing Value-Value (VV) attention and integrating CLIP ViT QKV weights, as shown in Figure \ref{fig:figure 2}c. The use of VV-attention is motivated by observations that the original QKV attention sometimes highlights features that are unintuitive to human perception, often identifying the background noise rather than semantically similar regions \cite{zhou_extract_2022, li_clip_2023}. Since anomaly segment annotations are human-labeled, we follow the CLIP surgery and adopt VV-attention to better align with human intuition. The attention weights are inherited from the corresponding layers of the CLIP visual encoder ($H_1$ to $H_4$).

\begin{figure}[t]
\centering
\includegraphics[width=0.8\columnwidth]{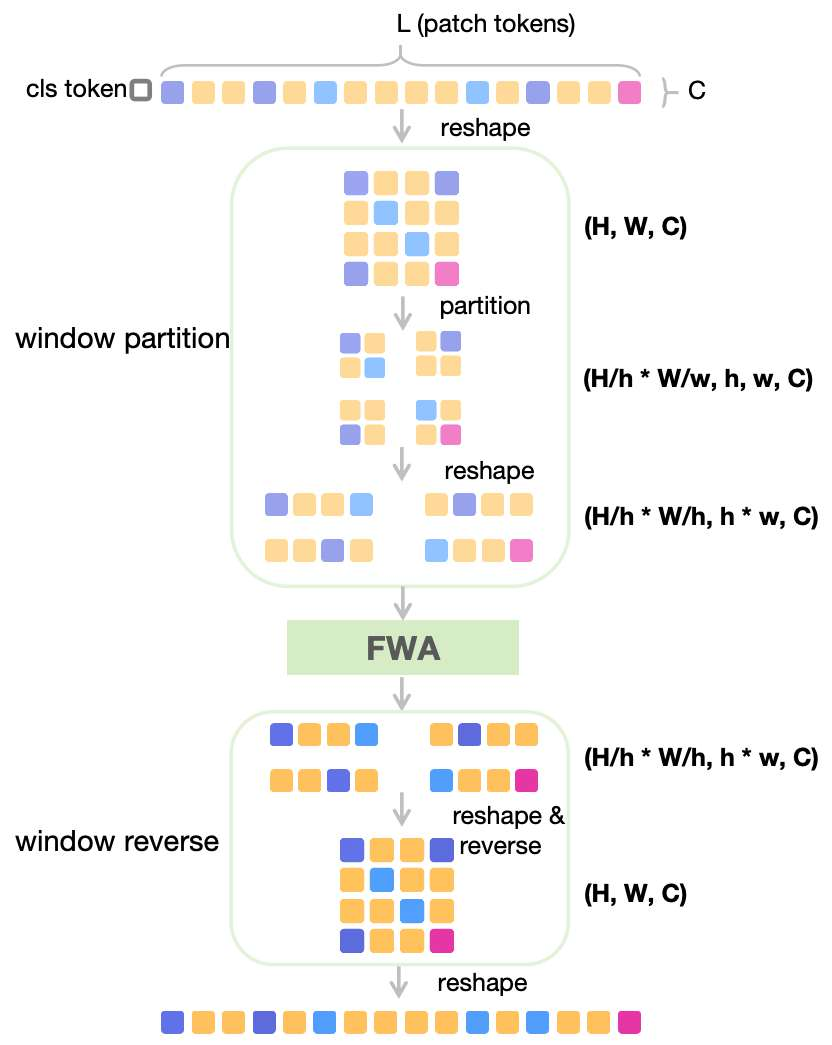}
\caption{The dimensional transformations of the input feature
through FWA adapter.}
\label{fig:figure 3}
\end{figure}

The dimensional transformations of the input feature through FWA adapter are detailed in Figure \ref{fig:figure 3}. Initially, the input features with dimensions $(L, C)$, where $L = H \times W$, are reshaped into a 2D feature map $(H, W, C)$. This map is divided into windows of size $h \times w$, resulting in $\frac{H}{h} \times \frac{W}{w}$ windows, each with dimensions $(h, w, C)$. After partitioning, the dimensions become $(\frac{H}{h} \times \frac{W}{w}, h, w, C)$. Next, each window is reshaped to $(h \times w, C)$ for attention computation, resulting in dimensions $(\frac{H}{h} \times \frac{W}{w}, h \times w, C)$. The VV-attention is applied within each window, maintaining these dimensions. Importantly, attention for all windows is computed in parallel for high efficiency. Subsequently, each window is reshaped back to $(h, w, C)$, giving overall dimensions of $(\frac{H}{h} \times \frac{W}{w}, h, w, C)$. The processed windows are recombined into the 2D feature map $(H, W, C)$, which is then reshaped back to the original 1D sequence $(L, C)$.

\subsection{Visual-Language Alignment}
\label{sec:align}

Anomalous regions are typically smaller than normal ones, so Focal Loss is used to address class imbalance \cite{lin_focal_2017}. Additionally, Dice loss is employed to ensure the model accurately defines decision boundaries by measuring overlap between predicted segmentation and ground truth \cite{milletari_v-net_2016}. This optimization uses a combined loss function:
\begin{equation}
\begin{aligned}
L = \quad &{\lambda}_1 \text{Dice}(\text{softmax}(F_{\text{fusion}}), S) + \\
&{\lambda}_2 \text{Focal}(\text{softmax}(F_{\text{fusion}}), S) + \\
&{\lambda}_3 \text{BCE}(\text{softmax}(F_{\text{cls}} F^T_{\text{text}}), c)
\end{aligned}
\tag{5}
\end{equation}
Here, $Dice(\cdot, \cdot)$, $Focal(\cdot, \cdot)$, and $BCE(\cdot, \cdot)$ denote the dice loss, focal loss, and binary cross-entropy loss, respectively. The weights ${\lambda}_1$, ${\lambda}_2$, and ${\lambda}_3$ are set to 1.0 by default.

\subsection{Few-Shot Inference}
\label{sec:few_shot}

\begin{figure}[t]
\centering
\includegraphics[width=\columnwidth]{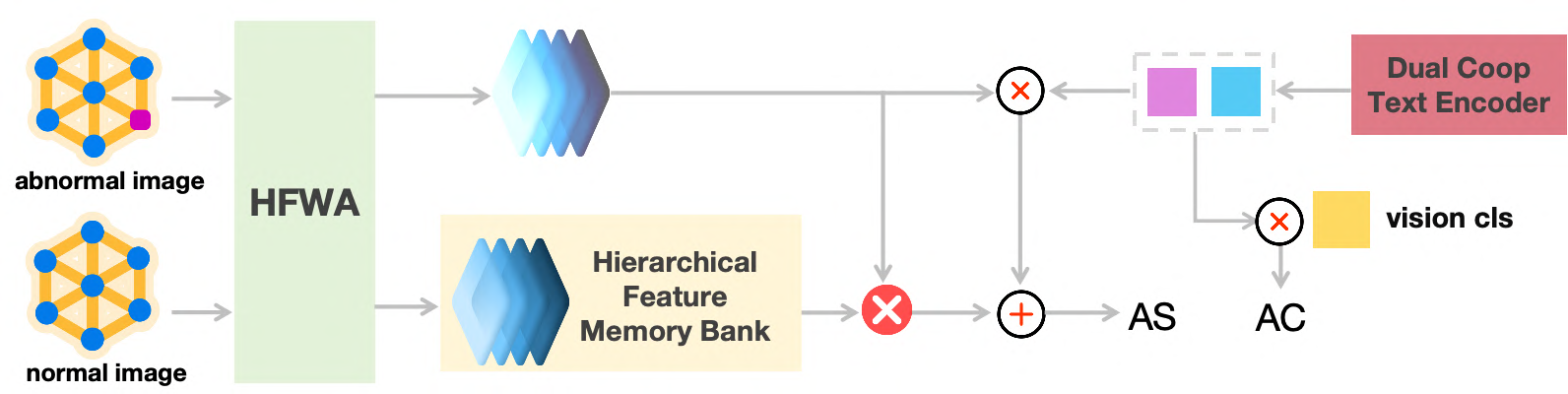}
\caption{Few-Shot inference framework.}
\label{fig:figure 4}
\end{figure}

Figure \ref{fig:figure 4} demonstrates the few-shot feature comparison method. Hierarchical features are extracted from normal images and stored in the feature memory bank \cite{roth_towards_2022, jeong_winclip_2023}. The features of the image to be inspected are then compared with these stored normal features to identify anomalous regions. This comparison enables the model to accurately locate and identify anomalies, enhancing the precision of anomaly detection. Detailed descriptions of few-shot inference are provided in Appendix A.

\section{Experiments}
\label{sec:experiments}

\subsection{Experimental Setups}

\noindent\textbf{Datasets.} In this paper, we utilize five benchmarks: MVTec-AD \cite{bergmann_mvtec_2019}, VisA \cite{zou_spot--difference_2022}, BTAD \cite{mishra_vt-adl_2021}, DAGM, and DTD-Synthetic \cite{aota_zero-shot_2023}. MVTec-AD includes 15 objects, with images having resolutions between 7002 and 9002 pixels, where the training set contains only normal samples, and the test set includes both normal and anomalous samples. The VisA dataset consists of 12 objects, with images approximately 1.5K × 1K pixels in size, following the same training and testing structure. BTAD is a real-world industrial dataset comprising 2830 images of three industrial products, focusing on body and surface defects. DAGM contains artificially generated images that simulate real-world problems, divided into ten subsets with 1000 defect-free and defect-labeled images. DTD-Synthetic includes synthesized anomalies in images ranging from 180 × 180 to 384 × 384 pixels, providing diverse textures and orientations samples. Details can be found in Appendix B.

\noindent\textbf{Competing Methods and Baselines.} We compare our HFWA model against competing models, WinCLIP and April-GAN, across five benchmark datasets. Additionally, we perform comparative experiments on the MVTec-AD and VisA datasets with several prominent methods: SPADE \cite{cohen_sub-image_2021}, PaDiM \cite{defard_padim_2021}, PatchCore \cite{roth_towards_2022}, PromptAD \cite{li_promptad_2024}, and InCTRL \cite{zhu_toward_2024}. SPADE, PaDiM, and PatchCore are well-established methods formulated for full-shot settings but adapted here for few-shot settings. Detailed information on these methods can be found in Appendix C.

\noindent\textbf{Evaluation Metrics.} The performance is quantified using the Area Under the Receiver Operating Characteristic Curve (AUROC) metric. This standard metric in anomaly detection evaluation considers image-level AUROC for anomaly classification (AC) and pixel-level AUROC for anomaly segmentation (AS). To enhance the evaluation, we use the Area Under the Average Precision (AP) and per-region-overlap (PRO) metrics \cite{gudovskiy_cflow-ad_2022}. PRO is particularly useful for precise anomaly localization, offering detailed performance assessment in scenarios like quality control where identifying exact defect areas is crucial \cite{bergmann_uninformed_2020}.

\noindent\textbf{Model and Training Details.} We use CLIP with ViT-L/14@336 as the backbone, dividing its 24 layers into 4 stages ($H_1$ to $H_4$) with 6 layers each. Learnable word embeddings are set to length 12. HFWA is fine-tuned on the MVTec-AD test data and assessed on other datasets. For assessing MVTec-AD, fine-tuning uses VisA test data. We employ the Adam optimizer with a learning rate of 1e-3, a batch size of 8, training for 1 epoch on an NVIDIA GeForce RTX 3070 8GB GPU.

\subsection{Main Results}
\label{sec:results}

\begin{table}[t]
\centering
\caption{Comparisons with competing few-shot anomaly detection methods on five datasets (K=4). The AUROC, AP for AC, AUROC and PRO for AS are reported. The best result is in \textbf{\textcolor{red}{red}}.}
\label{table:table1}
\resizebox{\columnwidth}{!}{%
\small 
\begin{tabular}{lcccc}
\toprule
\textbf{Metric} & \textbf{Dataset} & \textbf{WinCLIP} & \textbf{April-GAN} & \textbf{Ours} \\
\midrule
\multirow{5}{*}{AC AUROC} & MVTec-AD & 95.2±1.3 & 92.8±0.2 & \textbf{\textcolor{red}{96.8±0.3}} \\
 & Visa & 87.3±1.8 & 92.6±0.4 & \textbf{\textcolor{red}{92.9±0.2}} \\
 & BTAD & 87.0±0.2 & 92.1±0.2 & \textbf{\textcolor{red}{94.8±0.2}} \\
 & DAGM & 93.8±0.2 & 96.2±1.1 & \textbf{\textcolor{red}{98.9±0.3}} \\
 & DTD-Synthetic & 98.1±0.2 & 98.5±0.1 & \textbf{\textcolor{red}{99.1±0.0}} \\
\midrule
\multirow{5}{*}{AC AP} & MVTec-AD & 97.3±0.6 & 96.3±0.1 & \textbf{\textcolor{red}{98.3±0.3}} \\
 & Visa & 88.8±1.8 & 94.5±0.3 & \textbf{\textcolor{red}{94.5±0.2}} \\
 & BTAD & 86.8±0.0 & 95.2±0.5 & \textbf{\textcolor{red}{95.5±0.7}} \\
 & DAGM & 83.8±1.1 & 86.7±4.5 & \textbf{\textcolor{red}{95.2±1.7}} \\
 & DTD-Synthetic & 99.1±0.1 & 99.4±0.0 & \textbf{\textcolor{red}{99.6±0.0}} \\
\midrule
\multirow{5}{*}{AS AUROC} & MVTec-AD & \textbf{\textcolor{red}{96.2±0.3}} & 95.9±0.0 & 95.7±0.1 \\
 & Visa & \textbf{\textcolor{red}{97.2±0.2}} & 96.2±0.0 & 97.1±0.0 \\
 & BTAD & 95.8±0.0 & 94.4±0.1 & \textbf{\textcolor{red}{97.1±0.0}} \\
 & DAGM & 93.8±0.1 & 88.9±0.4 & \textbf{\textcolor{red}{96.9±0.0}} \\
 & DTD-Synthetic & 96.8±0.2 & 96.7±0.0 & \textbf{\textcolor{red}{98.7±0.0}} \\
\midrule
\multirow{5}{*}{AS AUPRO} & MVTec-AD & 89.0±0.8 & 91.8±0.1 & \textbf{\textcolor{red}{92.4±0.2}} \\
 & Visa & 87.6±0.9 & 90.2±0.1 & \textbf{\textcolor{red}{91.4±0.0}} \\
 & BTAD & 66.6±0.2 & 78.2±0.1 & \textbf{\textcolor{red}{81.2±0.2}} \\
 & DAGM & 82.4±0.3 & 77.8±0.9 & \textbf{\textcolor{red}{94.4±0.1}} \\
 & DTD-Synthetic & 90.1±0.5 & 92.2±0.0 & \textbf{\textcolor{red}{96.6±0.1}} \\
\bottomrule
\end{tabular}
}
\end{table}

\begin{figure}[t]
\centering
\includegraphics[width=\columnwidth]{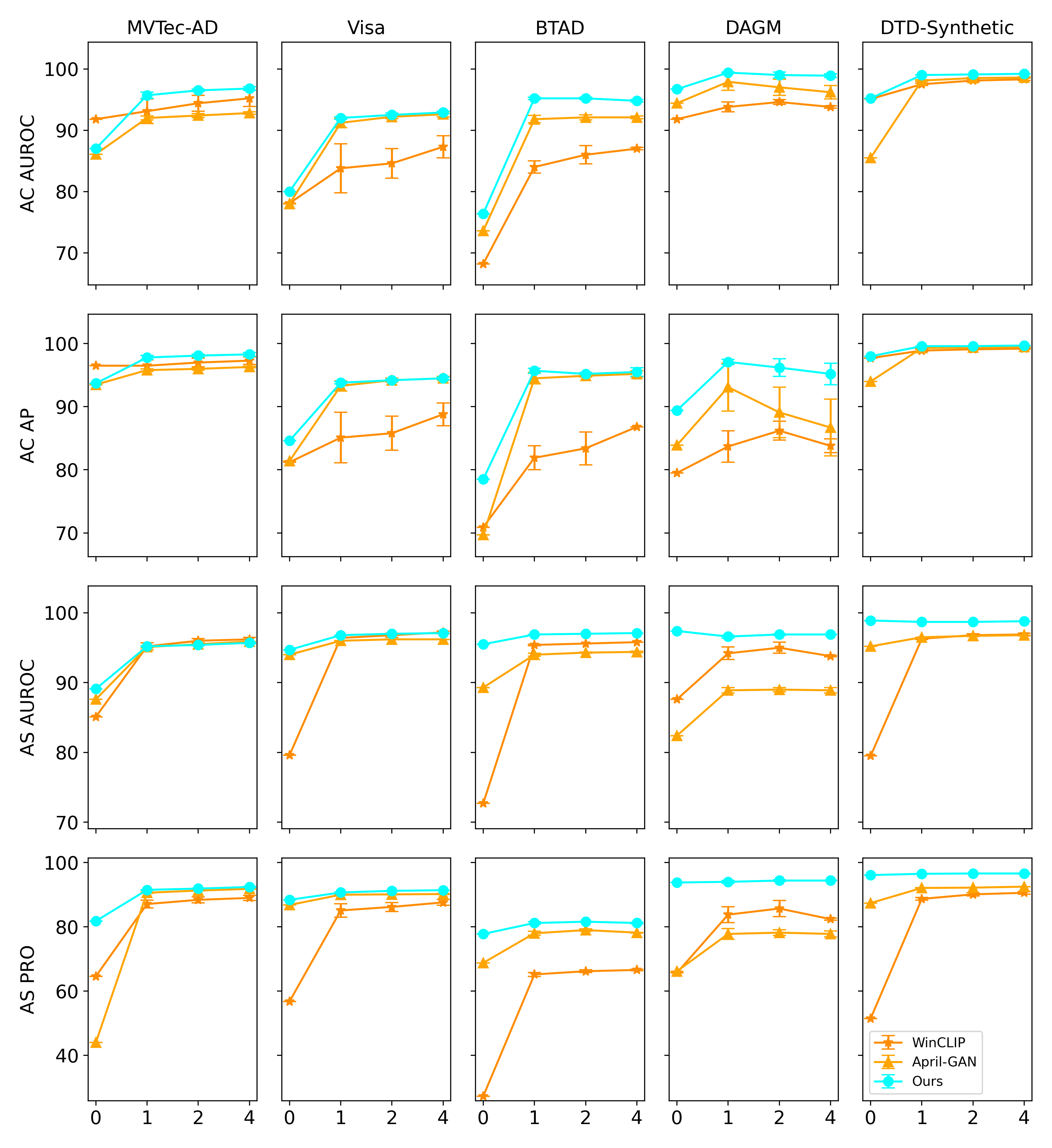}
\caption{Comparisons with zero-/few-shot anomaly detection methods on datasets of MVTec-AD, Visa, BTAD, DAGM and DTD Synthetic.}
\label{fig:figure 5}
\end{figure}

\begin{table}[t]
\centering
\caption{Performance Comparison on MVTec-AD and Visa Datasets. The best performance is highlighted in \textbf{\textcolor{red}{red}}, and performances lower than HFWA are shown in \textcolor{blue}{blue}.}
\label{table:table2}
\renewcommand{\arraystretch}{1.3} 
\small 
\resizebox{\columnwidth}{!}{%
\begin{tabular}{
    >{\centering\arraybackslash}m{1.8cm}
    >{\centering\arraybackslash}m{1.2cm}
    >{\centering\arraybackslash}m{1.2cm}
    >{\centering\arraybackslash}m{1.2cm}
    >{\centering\arraybackslash}m{1.2cm}
    >{\centering\arraybackslash}m{1.2cm}
    >{\centering\arraybackslash}m{1.2cm}}
\toprule
\multirow{2}{*}{\centering \shortstack{\textbf{Method} \\ \textbf{(Source)}}} & \multicolumn{3}{c}{\textbf{MVTec-AD}} & \multicolumn{3}{c}{\textbf{Visa}} \\
\cmidrule(lr){2-4} \cmidrule(lr){5-7}
 & \textbf{AC AUROC} & \textbf{AS AUROC} & \textbf{AS PRO} & \textbf{AC AUROC} & \textbf{AS AUROC} & \textbf{AS PRO} \\
\midrule
SPADE \newline (arXiv 2020) & \textcolor{blue}{84.8±2.5} & \textcolor{blue}{92.7±0.3} & \textcolor{blue}{87.0±0.5} & \textcolor{blue}{81.7±3.4} & \textcolor{blue}{96.6±0.3} & \textcolor{blue}{87.3±0.8} \\
PaDiM \newline (ICPR 2021) & \textcolor{blue}{80.4±2.4} & \textcolor{blue}{92.6±0.7} & \textcolor{blue}{81.3±1.9} & \textcolor{blue}{72.8±2.9} & \textcolor{blue}{93.2±0.5} & \textcolor{blue}{72.6±1.9} \\
PatchCore \newline (CVPR 2022) & \textcolor{blue}{88.8±2.6} & \textcolor{blue}{94.3±0.5} & \textcolor{blue}{84.3±1.6} & \textcolor{blue}{85.3±2.1} & \textcolor{blue}{96.8±0.3} & \textcolor{blue}{84.9±1.4} \\
WinCLIP \newline (CVPR 2023) & \textcolor{blue}{95.2±1.3} & 96.2±0.3 & \textcolor{blue}{89.0±0.8} & \textcolor{blue}{87.3±1.8} & 97.2±0.2 & \textcolor{blue}{87.6±0.9} \\
April-GAN \newline (arXiv 2023) & \textcolor{blue}{92.8±0.2} & 95.9±0.0 & \textcolor{blue}{91.8±0.1} & \textcolor{blue}{92.6±0.4} & \textcolor{blue}{96.2±0.0} & \textcolor{blue}{90.2±0.1} \\
PromptAD \newline (CVPR 2024) & \textcolor{blue}{96.6±0.9} & \textbf{\textcolor{red}{96.5±0.2}} & - & \textcolor{blue}{89.1±1.7} & \textbf{\textcolor{red}{97.4±0.3}} & - \\
InCTRL (CVPR 2024) & \textcolor{blue}{94.5±1.8} & - & - & \textcolor{blue}{87.7±1.9} & - & - \\
HFWA \newline (Ours) & \textbf{\textcolor{red}{96.8±0.3}} & 95.7±0.1 & \textbf{\textcolor{red}{92.4±0.2}} & \textbf{\textcolor{red}{92.9±0.2}} & 97.1±0.0 & \textbf{\textcolor{red}{91.4±0.0}} \\
\bottomrule
\end{tabular}
}
\end{table}

\noindent\textbf{Quantitative Results.} The results, as shown in Table \ref{table:table1}, demonstrate that our proposed method consistently outperforms WinCLIP and April-GAN across multiple datasets and evaluation metrics. Our method consistently demonstrated superior performance across the datasets. On the MVTec-AD dataset, our method achieved the highest AC AUROC score of 96.8±0.3, surpassing WinCLIP by 1.6 percentage points and April-GAN by 4.0 percentage points. Similarly, in the Visa dataset, our method achieved an AC AUROC score of 92.9±0.2, exceeding WinCLIP by 5.6 percentage points and April-GAN by 0.3 percentage points. On the BTAD dataset, our method reached 94.8±0.2, which is 7.8 percentage points higher than WinCLIP and 2.7 percentage points higher than April-GAN. In the DAGM dataset, our method achieved an impressive score of 99.1, outperforming WinCLIP by 5.3 percentage points and April-GAN by 4.5 percentage points. Lastly, on the DTD-Synthetic dataset, our method scored 99.1±0.0, which is 1.0 percentage point higher than WinCLIP and 0.6 percentage points higher than April-GAN. These results highlight the robustness and accuracy of our approach in image-level anomaly detection.

In terms of pixel-level performance, our method also showed significant improvements. Across all datasets, our method demonstrated superior performance in AS PRO, consistently leading over both WinCLIP and April-GAN. For instance, on the DAGM dataset, our method achieved an exceptional AS PRO score of 96.6, surpassing WinCLIP by 14.2 percentage points and April-GAN by 20.1 percentage points. This demonstrates our method's effectiveness in detecting fine-grained anomalies at the pixel level.

An in-depth analysis of performance across various zero- and few-shot scenarios ($K \in \{0, 1, 2, 4\}$), as depicted in Figure \ref{fig:figure 5}, indicates that our method consistently maintains superior performance across various datasets and metrics compared to WinCLIP and April-GAN. Specifically, our method achieves higher scores more consistently, demonstrating robustness and reliability in few-shot scenarios. Additionally, Table \ref{table:table2} shows a comparison with existing many-shot methods in AUROC (image and pixel level) and PRO on MVTec-AD and Visa. Our method consistently achieved higher scores, further validating its effectiveness and efficiency in anomaly detection tasks even when compared to well-established many-shot methods.

However, it is worth noting that our method fell slightly behind in AS AUROC on the MVTec-AD and Visa datasets compared to WinCLIP. The reasons for this discrepancy will be further analyzed in Sec. \ref{sec:performance}.

\begin{figure}[t]
\centering
\includegraphics[width=0.9\columnwidth]{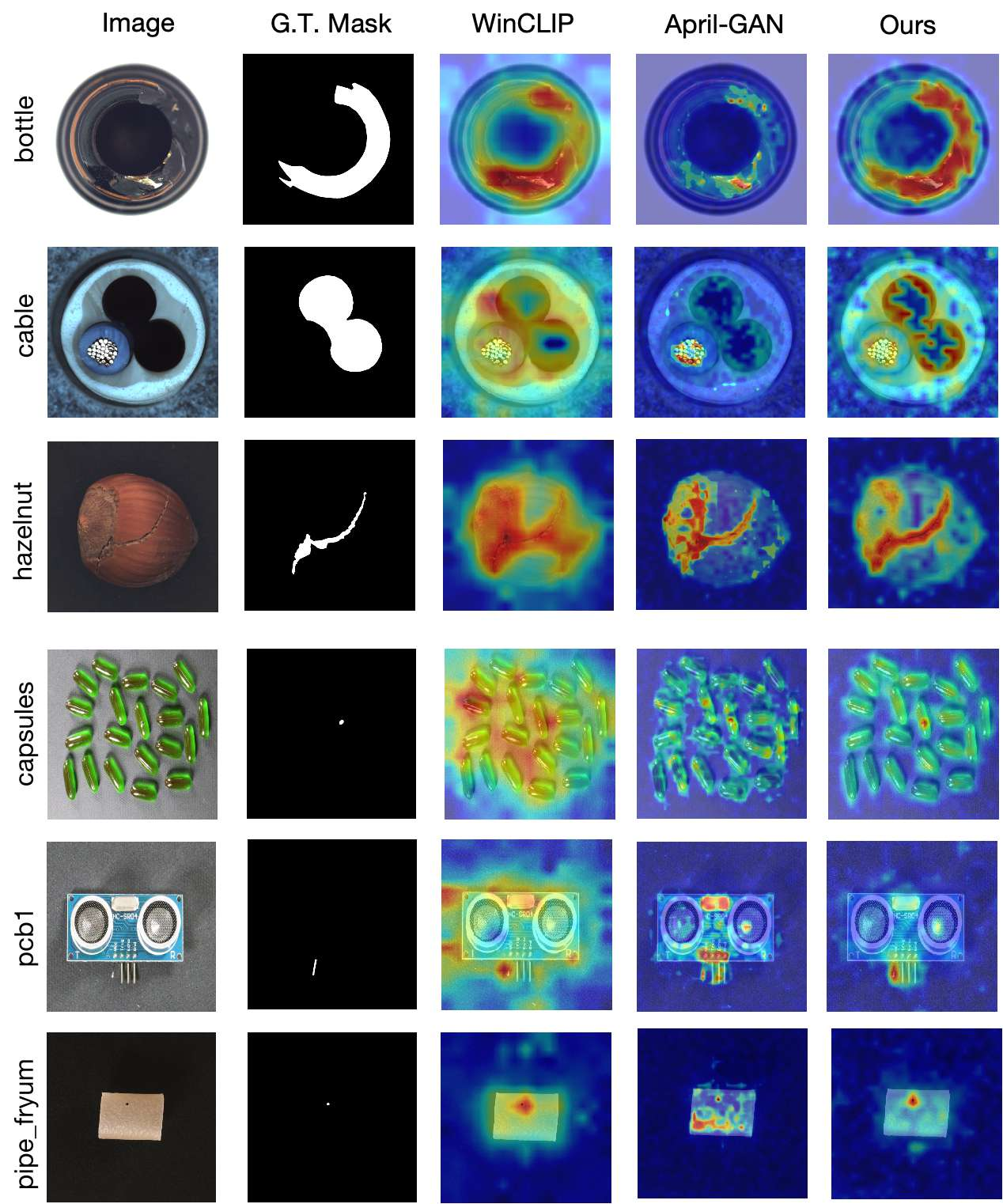}
\caption{Visualization results under the few-shot setting (K=4).}
\label{fig:figure 6}
\end{figure}

\noindent\textbf{Analysis of Anomaly Detection Results.} A comparative visualization of anomaly detections on MVTec-AD and Visa by WinCLIP, April-GAN, and HFWA has been shown in Figure \ref{fig:figure 6}. The visual analysis reveals distinct characteristics of each method: WinCLIP has moderate ability with obvious dispersed heatmaps, struggling with fine-grained anomalies; April-GAN improves localization but lacks precision for small defects, often producing false positives in complicated situations. Our method excels with precise, concentrated heatmaps that match the ground truth, demonstrating superior accuracy and robustness across various objects and defect types.

\subsection{Mechanism Analysis}
\label{sec:mechanism}

\begin{figure}[t]
\centering
\includegraphics[width=0.9\columnwidth]{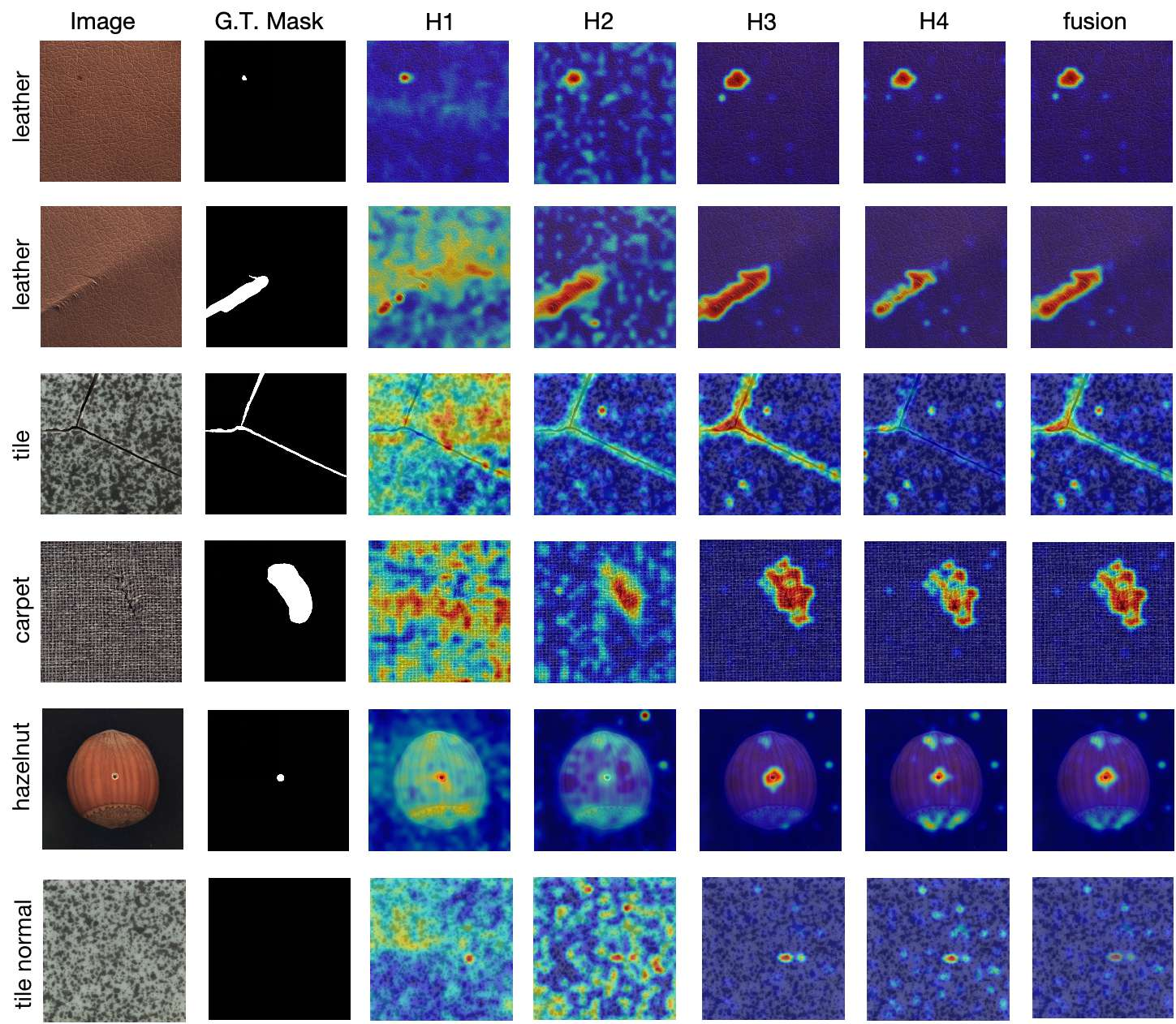}
\caption{Hierarchical Results on MVTec-AD Dataset. A set of images showing the real outputs of the model, illustrating how different layers (H1 to H4) process various feature modes. Each row represents a different sample, with columns showing the original image, segmentation mask, heatmap, and feature outputs from H1 to H4, and fusion.}
\label{fig:figure 7}
\end{figure}

Figure \ref{fig:figure 1} illustrates the proposed model's handling of different image feature patterns (point, line, plane, motley) across various layers (H1 to H4), with the number of stars indicating the model's depth. Figure \ref{fig:figure 7} presents the actual results on the MVTec-AD dataset, showing how these layers process different feature modes.

\noindent\textbf{Point Pattern.} Leather and Hazelnut with Point Defects (Row 1 and Raw 5): In the shallow layer H1 (point mode, 1-star), small point defects are accurately detected, the point defects are clear and precisely located. In H2 (point mode, 2-star), defect areas expand due to the attention mechanism but remain clear. In deeper layers H3-4 (point mode, 3-star), further attention spread reduces defect clarity, the point area becomes slightly contracted.

\noindent\textbf{Line and Motley Pattern.} Tile and Fabric with Cracks (Row 3 and Row 4): In the shallow layer H1 (motley and line mode, 1-star), only local normal texture is detected. Normal texture is evident, but cracks are not visible. In middle layers H2-3, cracks are captured by balancing local and global features, making them visible. In the deepest layer H4, wide attention reduces the clarity of crack features. In particular, cracks in Tile nearly disappear due to the narrowness of local features.

\noindent\textbf{Mixed Point and Line Pattern.} Leather with Fold Defects (Row 2): In the shallow layer H1 (point and line mode, 1-star), very localized defects are detected. In the middle layer H2, the model starts to understand longer-range line features, making the defects clearer over a broader area. In deeper layers H3, the clarity of these features continues to improve, showing both local and extended defects more distinctly. In deeper layers H4, the abnormal area shrinks as more normal features are attended to. In this case, the abnormal features do not disappear as tile cracks due to the greater local depth.

\noindent\textbf{Discussion.} The analysis confirms that the results in Figure \ref{fig:figure 7} correspond to the theoretical model illustrated in Figure \ref{fig:figure 1}. Point features are most precise in the shallow layer (H1), begin to spread in the middle layer (H2), and further spread in the deep layers (H3-4), reducing their distinctness. Line and motley features initially show normal texture in the shallow layer (H1), and become clearer in the middle layers (H2-3), but their distinctness diminishes in the deepest layer (H4) due to wide attention. This hierarchical feature extraction and aggregation method demonstrates the model's capability to handle different image feature modes at various depths, confirming its effectiveness in complex image scenarios.

\subsection{Performance Analysis}
\label{sec:performance}

\begin{figure}[t]
\centering
\includegraphics[width=0.95\columnwidth]{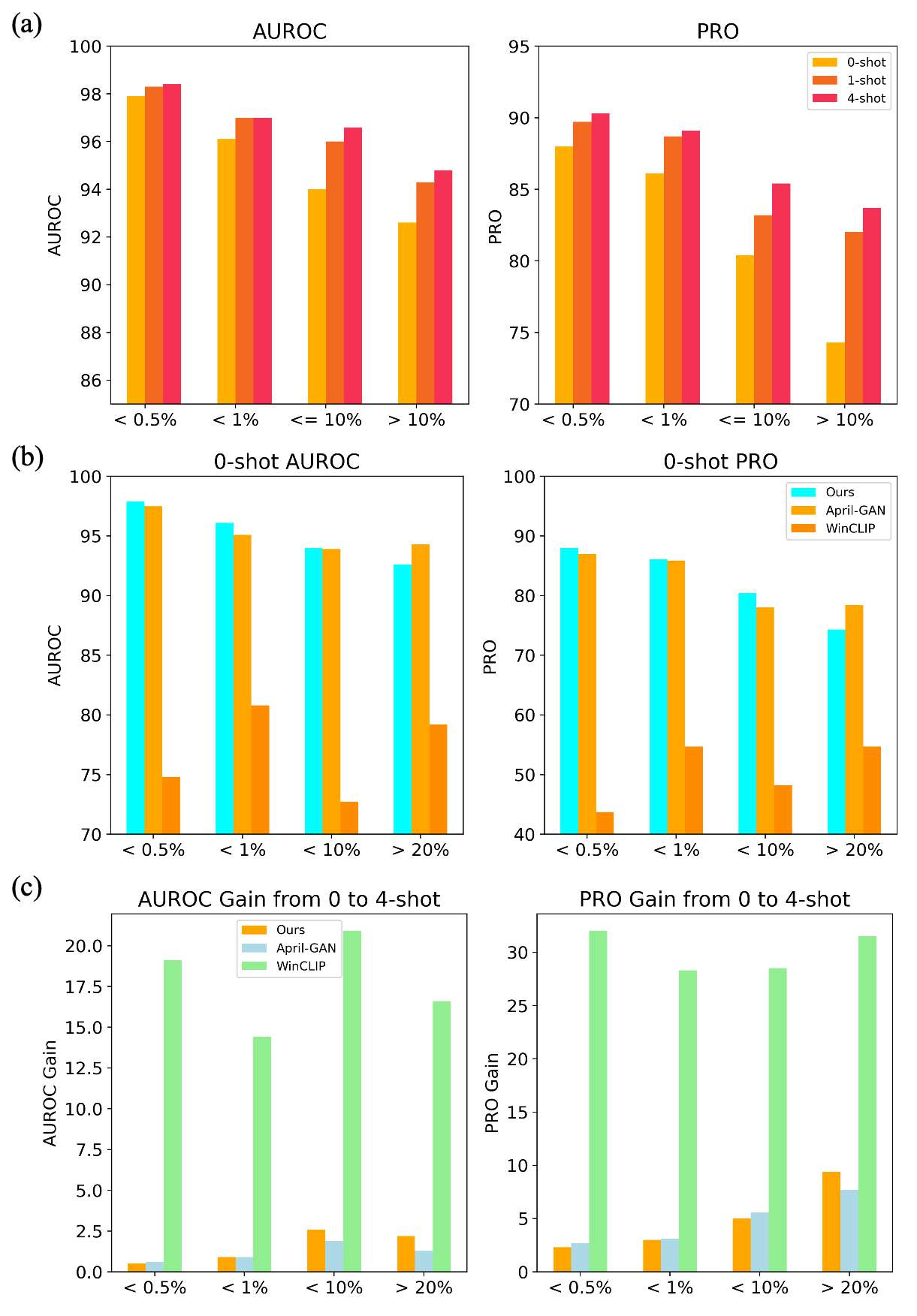}
\caption{(a) Performance comparison across different shot settings; (b) Zero-shot performance comparison with baselines; (c) Performance gain from 0-shot to 4-shot of different methods.}
\label{fig:figure 8}
\end{figure}

To understand the model's performance, we divided the Visa dataset into four subsets based on the proportion of the anomalous region: $<0.5\%$ (920 samples), $<1\%$ (87 samples), $<=10\%$ (156 samples), and $>10\%$ (37 samples). We then studied the performance of different models on these subsets and compared the zero-shot capabilities with the performance gains achieved through few-shot learning. Figure \ref{fig:figure 8}a shows the AUROC and PRO metrics for zero/few-shot settings across various anomaly sizes. The results indicate that as shots increase from 0 to 4, the metrics generally improve, with a notable increase for anomalies greater than 10\%. This suggests that few-shot learning significantly enhances the model's ability to detect larger anomalies. Figure \ref{fig:figure 8}b compares the zero-shot metrics of our method with April-GAN and WinCLIP. Our method consistently outperforms April-GAN for anomalies less than 10\%, for anomalies greater than 10\%, April-GAN performs better. WinCLIP consistently underperforms compared to both our method and April-GAN across all anomaly sizes. However, WinCLIP shows more significant gains than our method and April-GAN across all anomaly sizes, as illustrated in Figure \ref{fig:figure 8}c, indicating the substantial difference in the mechanisms of non-fine-tuned and fine-tuned methods.

The results collectively demonstrate that our method outperforms existing approaches in zero-shot settings for smaller anomalies ($<=10\%$) but is surpassed by WinCLIP for larger anomalies ($>10\%$). Few-shot learning significantly enhances the performance of all models, with WinCLIP benefiting the most across all anomaly sizes.

\begin{table}[t]
\centering
\caption{Comparison of AUROC and PRO values across different samples and methods.}
\label{table:table3}
\renewcommand{\arraystretch}{1.3} 
\small 
\resizebox{\columnwidth}{!}{%
\begin{tabular}{
    >{\centering\arraybackslash}m{1.8cm} 
    >{\centering\arraybackslash}m{1.2cm} 
    >{\centering\arraybackslash}m{1.2cm} 
    >{\centering\arraybackslash}m{1.2cm} 
    >{\centering\arraybackslash}m{1.2cm} 
    >{\centering\arraybackslash}m{1.2cm} 
    >{\centering\arraybackslash}m{1.2cm}}
\toprule
\multirow{2}{*}{\textbf{Samples}} & \multicolumn{2}{c}{\textbf{Ours}} & \multicolumn{2}{c}{\textbf{April-GAN}} & \multicolumn{2}{c}{\textbf{WinCLIP}} \\
\cmidrule(lr){2-3} \cmidrule(lr){4-5} \cmidrule(lr){6-7}
& \textbf{AUROC} & \textbf{PRO} & \textbf{AUROC} & \textbf{PRO} & \textbf{AUROC} & \textbf{PRO} \\
\midrule
pcb1 \newline (no) & \textbf{\textcolor{red}{96.4}} & \textbf{\textcolor{red}{88.4}} & 95.2 & 85.3 & 95.5 & 85.2 \\
fryum \newline (partial) & 96.3 & 87.5 & \textbf{\textcolor{red}{97}} & \textbf{\textcolor{red}{89.1}} & 95.8 & 85.8 \\
cashew \newline (all) & 90.1 & 69.9 & 91.6 & 74.5 & \textbf{\textcolor{red}{94}} & \textbf{\textcolor{red}{79.4}} \\
pipe\_fryum  \newline (all) & 96.4 & 88.8 & 95.2 & \textbf{\textcolor{red}{95.5}} & \textbf{\textcolor{red}{98.1}} & 94.2 \\
\bottomrule
\end{tabular}
}
\end{table}

Further, to detail the reasons why the performance drops when anomaly size exceeds $10\%$, we manually examined the 37 samples, finding that our method only underperforms compared to others in cases of polyploidy. The results are shown in Table \ref{table:table3}. Our method typically constructs anomaly features from local to global patterns. However, since polyploidy usually results from the overlay of normal objects, it lacks rule-based semantic definitions, leading to a decline in our performance. The detailed analysis is included in Appendix D.

\subsection{Inference Speed}
\label{sec:inference}

\begin{figure}[t]
\centering
\includegraphics[width=\columnwidth]{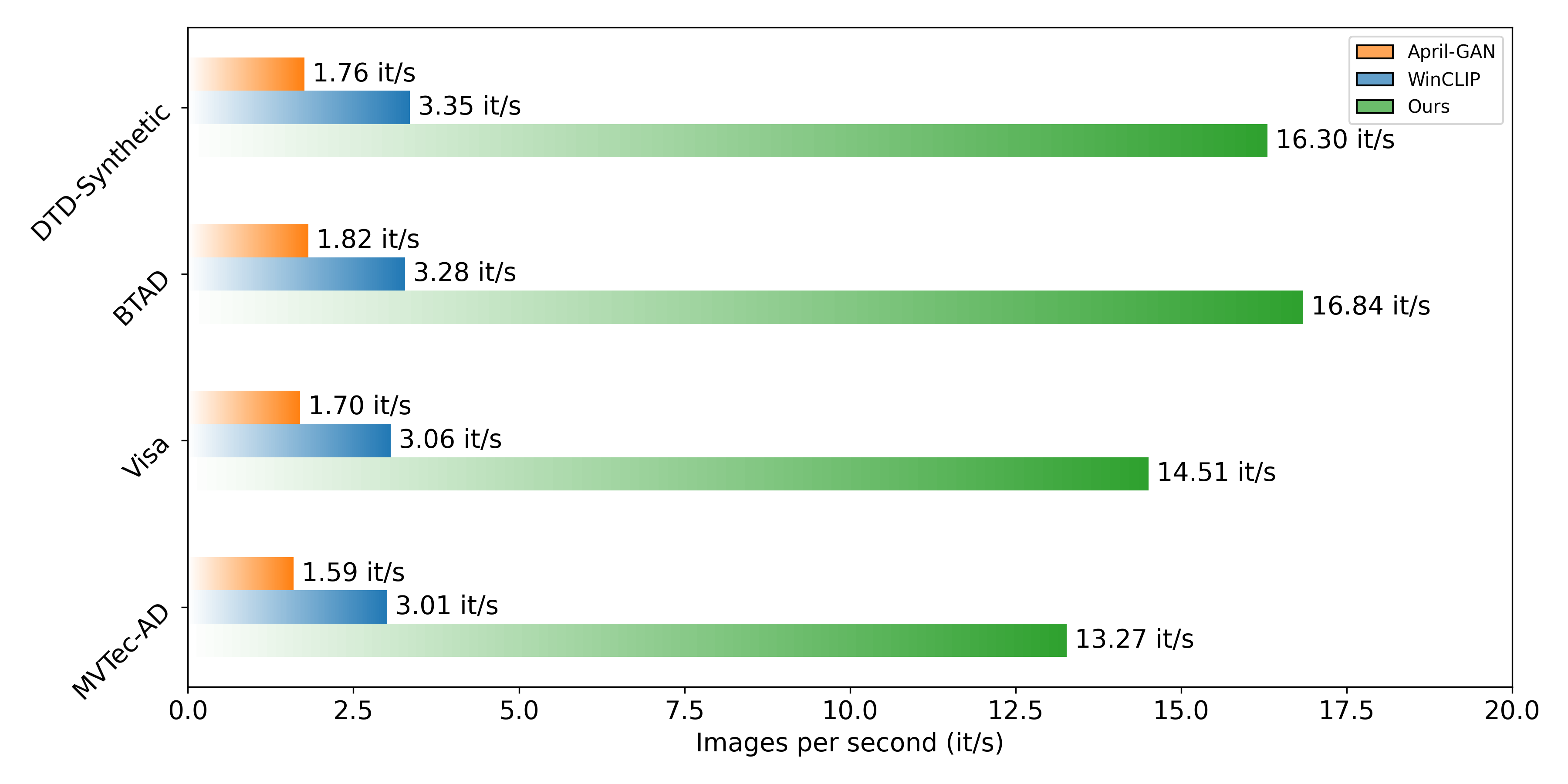}
\caption{Inference performance comparison of different methods on a single NVIDIA RTX3070 8GB GPU.}
\label{fig:figure 9}
\end{figure}

Our method network maintains high inference speed. As shown in Figure \ref{fig:figure 9}, our model is more efficient than April-GAN and WinCLIP. For instance, on the BTAD dataset, our model processes at 16.84 it/s, compared to WinCLIP's 3.28 it/s and April-GAN's 1.82 it/s. This efficiency is due to windowed self-attention and frozen weights.

\section{Conclusion}

This paper presents an elaborately designed framework based on vision-language models for anomaly detection. By incorporating a window self-attention mechanism and learnable prompts, our approach adaptively focuses on relevant features at various hierarchical levels, improving anomaly detection accuracy. Experiments show that this method outperforms current state-of-the-art techniques across multiple benchmark datasets, significantly improving anomaly classification and segmentation tasks. Our research validates the effectiveness of the model and offers valuable directions for future research in this field.


{\small
\bibliographystyle{ieee_fullname}
\bibliography{mylibrary}

\begin{thebibliography}{10}\itemsep=-1pt

\bibitem{aota_zero-shot_2023}
Toshimichi Aota, Lloyd Teh~Tzer Tong, and Takayuki Okatani.
\newblock Zero-shot versus many-shot: Unsupervised texture anomaly detection.
\newblock In {\em 2023 {IEEE}/{CVF} Winter Conference on Applications of Computer Vision ({WACV})}, pages 5553--5561, 2023.
\newblock {ISSN}: 2642-9381.

\bibitem{bae_pni_2023}
Jaehyeok Bae, Jae-Han Lee, and Seyun Kim.
\newblock {PNI}: Industrial anomaly detection using position and neighborhood information.
\newblock In {\em 2023 {IEEE}/{CVF} International Conference on Computer Vision ({ICCV})}, pages 6350--6360, 2023.
\newblock {ISSN}: 2380-7504.

\bibitem{batzner_efficientad_2024}
Kilian Batzner, Lars Heckler, and Rebecca König.
\newblock {EfficientAD}: Accurate visual anomaly detection at millisecond-level latencies.
\newblock In {\em 2024 {IEEE}/{CVF} Winter Conference on Applications of Computer Vision ({WACV})}, pages 127--137, 2024.
\newblock {ISSN}: 2642-9381.

\bibitem{bergmann_beyond_2022}
Paul Bergmann, Kilian Batzner, Michael Fauser, David Sattlegger, and Carsten Steger.
\newblock Beyond dents and scratches: Logical constraints in unsupervised anomaly detection and localization.
\newblock 130(4):947--969, 2022.

\bibitem{bergmann_mvtec_2019}
Paul Bergmann, Michael Fauser, David Sattlegger, and Carsten Steger.
\newblock {MVTec} {AD} — a comprehensive real-world dataset for unsupervised anomaly detection.
\newblock In {\em 2019 {IEEE}/{CVF} Conference on Computer Vision and Pattern Recognition ({CVPR})}, pages 9584--9592, 2019.
\newblock {ISSN}: 2575-7075.

\bibitem{bergmann_uninformed_2020}
Paul Bergmann, Michael Fauser, David Sattlegger, and Carsten Steger.
\newblock Uninformed students: Student-teacher anomaly detection with discriminative latent embeddings.
\newblock In {\em 2020 {IEEE}/{CVF} Conference on Computer Vision and Pattern Recognition ({CVPR})}, pages 4182--4191, 2020.
\newblock {ISSN}: 2575-7075.

\bibitem{chen_simple_2020}
Ting Chen, Simon Kornblith, Mohammad Norouzi, and Geoffrey Hinton.
\newblock A simple framework for contrastive learning of visual representations.
\newblock In {\em Proceedings of the 37th International Conference on Machine Learning}, pages 1597--1607. {PMLR}, 2020.
\newblock {ISSN}: 2640-3498.

\bibitem{chen_april-gan_2023}
Xuhai Chen, Yue Han, and Jiangning Zhang.
\newblock {APRIL}-{GAN}: A zero-/few-shot anomaly classification and segmentation method for {CVPR} 2023 {VAND} workshop challenge tracks 1\&2: 1st place on zero-shot {AD} and 4th place on few-shot {AD}, 2023.

\bibitem{cohen_sub-image_2021}
Niv Cohen and Yedid Hoshen.
\newblock Sub-image anomaly detection with deep pyramid correspondences, 2021.

\bibitem{defard_padim_2021}
Thomas Defard, Aleksandr Setkov, Angelique Loesch, and Romaric Audigier.
\newblock {PaDiM}: A patch distribution modeling framework for anomaly detection and localization.
\newblock In Alberto Del~Bimbo, Rita Cucchiara, Stan Sclaroff, Giovanni~Maria Farinella, Tao Mei, Marco Bertini, Hugo~Jair Escalante, and Roberto Vezzani, editors, {\em Pattern Recognition. {ICPR} International Workshops and Challenges}, pages 475--489. Springer International Publishing, 2021.

\bibitem{deng_anomaly_2022}
Hanqiu Deng and Xingyu Li.
\newblock Anomaly detection via reverse distillation from one-class embedding.
\newblock In {\em 2022 {IEEE}/{CVF} Conference on Computer Vision and Pattern Recognition ({CVPR})}, pages 9727--9736, 2022.
\newblock {ISSN}: 2575-7075.

\bibitem{dosovitskiy_image_2020}
Alexey Dosovitskiy, Lucas Beyer, Alexander Kolesnikov, Dirk Weissenborn, Xiaohua Zhai, Thomas Unterthiner, Mostafa Dehghani, Matthias Minderer, Georg Heigold, Sylvain Gelly, Jakob Uszkoreit, and Neil Houlsby.
\newblock An image is worth 16x16 words: Transformers for image recognition at scale.
\newblock 2020.

\bibitem{gu_open-vocabulary_2021}
Xiuye Gu, Tsung-Yi Lin, Weicheng Kuo, and Yin Cui.
\newblock Open-vocabulary object detection via vision and language knowledge distillation.
\newblock 2021.

\bibitem{gudovskiy_cflow-ad_2022}
Denis Gudovskiy, Shun Ishizaka, and Kazuki Kozuka.
\newblock {CFLOW}-{AD}: Real-time unsupervised anomaly detection with localization via conditional normalizing flows.
\newblock In {\em 2022 {IEEE}/{CVF} Winter Conference on Applications of Computer Vision ({WACV})}, pages 1819--1828, 2022.
\newblock {ISSN}: 2642-9381.

\bibitem{huang_registration_2022}
Chaoqin Huang, Haoyan Guan, Aofan Jiang, Ya Zhang, Michael Spratling, and Yan-Feng Wang.
\newblock Registration based few-shot anomaly detection.
\newblock In Shai Avidan, Gabriel Brostow, Moustapha Cissé, Giovanni~Maria Farinella, and Tal Hassner, editors, {\em Computer Vision – {ECCV} 2022}, pages 303--319. Springer Nature Switzerland, 2022.

\bibitem{jeong_winclip_2023}
Jongheon Jeong, Yang Zou, Taewan Kim, Dongqing Zhang, Avinash Ravichandran, and Onkar Dabeer.
\newblock {WinCLIP}: Zero-/few-shot anomaly classification and segmentation.
\newblock pages 19606--19616, 2023.

\bibitem{jia_scaling_2021}
Chao Jia, Yinfei Yang, Ye Xia, Yi-Ting Chen, Zarana Parekh, Hieu Pham, Quoc Le, Yun-Hsuan Sung, Zhen Li, and Tom Duerig.
\newblock Scaling up visual and vision-language representation learning with noisy text supervision.
\newblock In {\em Proceedings of the 38th International Conference on Machine Learning}, pages 4904--4916. {PMLR}, 2021.
\newblock {ISSN}: 2640-3498.

\bibitem{jiang_how_2020}
Zhengbao Jiang, Frank~F. Xu, Jun Araki, and Graham Neubig.
\newblock How can we know what language models know?
\newblock 8:423--438, 2020.

\bibitem{li_cutpaste_2021}
Chun-Liang Li, Kihyuk Sohn, Jinsung Yoon, and Tomas Pfister.
\newblock {CutPaste}: Self-supervised learning for anomaly detection and localization.
\newblock In {\em 2021 {IEEE}/{CVF} Conference on Computer Vision and Pattern Recognition ({CVPR})}, pages 9659--9669, 2021.
\newblock {ISSN}: 2575-7075.

\bibitem{li_promptad_2024}
Xiaofan Li, Zhizhong Zhang, Xin Tan, Chengwei Chen, Yanyun Qu, Yuan Xie, and Lizhuang Ma.
\newblock {PromptAD}: Learning prompts with only normal samples for few-shot anomaly detection, 2024.

\bibitem{li_prefix-tuning_2021}
Xiang~Lisa Li and Percy Liang.
\newblock Prefix-tuning: Optimizing continuous prompts for generation.
\newblock In Chengqing Zong, Fei Xia, Wenjie Li, and Roberto Navigli, editors, {\em Proceedings of the 59th Annual Meeting of the Association for Computational Linguistics and the 11th International Joint Conference on Natural Language Processing (Volume 1: Long Papers)}, pages 4582--4597. Association for Computational Linguistics, 2021.

\bibitem{li_clip_2023}
Yi Li, Hualiang Wang, Yiqun Duan, and Xiaomeng Li.
\newblock {CLIP} surgery for better explainability with enhancement in open-vocabulary tasks, 2023.

\bibitem{lin_focal_2017}
Tsung-Yi Lin, Priya Goyal, Ross Girshick, Kaiming He, and Piotr Dollár.
\newblock Focal loss for dense object detection.
\newblock In {\em 2017 {IEEE} International Conference on Computer Vision ({ICCV})}, pages 2999--3007, 2017.
\newblock {ISSN}: 2380-7504.

\bibitem{liu_simplenet_2023}
Zhikang Liu, Yiming Zhou, Yuansheng Xu, and Zilei Wang.
\newblock {SimpleNet}: A simple network for image anomaly detection and localization.
\newblock In {\em 2023 {IEEE}/{CVF} Conference on Computer Vision and Pattern Recognition ({CVPR})}, pages 20402--20411, 2023.
\newblock {ISSN}: 2575-7075.

\bibitem{milletari_v-net_2016}
Fausto Milletari, Nassir Navab, and Seyed-Ahmad Ahmadi.
\newblock V-net: Fully convolutional neural networks for volumetric medical image segmentation.
\newblock In {\em 2016 Fourth International Conference on 3D Vision (3DV)}, pages 565--571, 2016.

\bibitem{mishra_vt-adl_2021}
Pankaj Mishra, Riccardo Verk, Daniele Fornasier, Claudio Piciarelli, and Gian~Luca Foresti.
\newblock {VT}-{ADL}: A vision transformer network for image anomaly detection and localization.
\newblock In {\em 2021 {IEEE} 30th International Symposium on Industrial Electronics ({ISIE})}, pages 01--06, 2021.
\newblock {ISSN}: 2163-5145.

\bibitem{radford_learning_2021}
Alec Radford, Jong~Wook Kim, Chris Hallacy, Aditya Ramesh, Gabriel Goh, Sandhini Agarwal, Girish Sastry, Amanda Askell, Pamela Mishkin, Jack Clark, Gretchen Krueger, and Ilya Sutskever.
\newblock Learning transferable visual models from natural language supervision.
\newblock In {\em Proceedings of the 38th International Conference on Machine Learning}, pages 8748--8763. {PMLR}, 2021.
\newblock {ISSN}: 2640-3498.

\bibitem{rao_denseclip_2022}
Yongming Rao, Wenliang Zhao, Guangyi Chen, Yansong Tang, Zheng Zhu, Guan Huang, Jie Zhou, and Jiwen Lu.
\newblock {DenseCLIP}: Language-guided dense prediction with context-aware prompting.
\newblock In {\em 2022 {IEEE}/{CVF} Conference on Computer Vision and Pattern Recognition ({CVPR})}, pages 18061--18070, 2022.
\newblock {ISSN}: 2575-7075.

\bibitem{ristea_self-supervised_2022}
Nicolae-Cătălin Ristea, Neelu Madan, Radu~Tudor Ionescu, Kamal Nasrollahi, Fahad~Shahbaz Khan, Thomas~B. Moeslund, and Mubarak Shah.
\newblock Self-supervised predictive convolutional attentive block for anomaly detection.
\newblock In {\em 2022 {IEEE}/{CVF} Conference on Computer Vision and Pattern Recognition ({CVPR})}, pages 13566--13576, 2022.
\newblock {ISSN}: 2575-7075.

\bibitem{roth_towards_2022}
Karsten Roth, Latha Pemula, Joaquin Zepeda, Bernhard Schölkopf, Thomas Brox, and Peter Gehler.
\newblock Towards total recall in industrial anomaly detection.
\newblock In {\em 2022 {IEEE}/{CVF} Conference on Computer Vision and Pattern Recognition ({CVPR})}, pages 14298--14308, 2022.
\newblock {ISSN}: 2575-7075.

\bibitem{ruff_deep_2018}
Lukas Ruff, Robert Vandermeulen, Nico Goernitz, Lucas Deecke, Shoaib~Ahmed Siddiqui, Alexander Binder, Emmanuel Müller, and Marius Kloft.
\newblock Deep one-class classification.
\newblock In {\em Proceedings of the 35th International Conference on Machine Learning}, pages 4393--4402. {PMLR}, 2018.
\newblock {ISSN}: 2640-3498.

\bibitem{sheynin_hierarchical_2021}
Shelly Sheynin, Sagie Benaim, and Lior Wolf.
\newblock A hierarchical transformation-discriminating generative model for few shot anomaly detection.
\newblock In {\em 2021 {IEEE}/{CVF} International Conference on Computer Vision ({ICCV})}, pages 8475--8484, 2021.
\newblock {ISSN}: 2380-7504.

\bibitem{shin_autoprompt_2020}
Taylor Shin, Yasaman Razeghi, Robert~L. Logan~{IV}, Eric Wallace, and Sameer Singh.
\newblock {AutoPrompt}: Eliciting knowledge from language models with automatically generated prompts.
\newblock In Bonnie Webber, Trevor Cohn, Yulan He, and Yang Liu, editors, {\em Proceedings of the 2020 Conference on Empirical Methods in Natural Language Processing ({EMNLP})}, pages 4222--4235. Association for Computational Linguistics, 2020.

\bibitem{sun_dualcoop_2022}
Ximeng Sun, Ping Hu, and Kate Saenko.
\newblock {DualCoOp}: Fast adaptation to multi-label recognition with limited annotations.
\newblock 35:30569--30582, 2022.

\bibitem{vaswani_attention_2017}
Ashish Vaswani, Noam Shazeer, Niki Parmar, Jakob Uszkoreit, Llion Jones, Aidan~N Gomez, Ł~ukasz Kaiser, and Illia Polosukhin.
\newblock Attention is all you need.
\newblock In {\em Advances in Neural Information Processing Systems}, volume~30. Curran Associates, Inc., 2017.

\bibitem{wyatt_anoddpm_2022}
Julian Wyatt, Adam Leach, Sebastian~M. Schmon, and Chris~G. Willcocks.
\newblock {AnoDDPM}: Anomaly detection with denoising diffusion probabilistic models using simplex noise.
\newblock In {\em 2022 {IEEE}/{CVF} Conference on Computer Vision and Pattern Recognition Workshops ({CVPRW})}, pages 649--655, 2022.
\newblock {ISSN}: 2160-7516.

\bibitem{xing_less_2024}
Yujie Xing, Xiao Wang, Yibo Li, Hai Huang, and Chuan Shi.
\newblock Less is more: on the over-globalizing problem in graph transformers, 2024.

\bibitem{yi_patch_2021}
Jihun Yi and Sungroh Yoon.
\newblock Patch {SVDD}: Patch-level {SVDD} for anomaly detection and segmentation.
\newblock In Hiroshi Ishikawa, Cheng-Lin Liu, Tomas Pajdla, and Jianbo Shi, editors, {\em Computer Vision – {ACCV} 2020}, pages 375--390. Springer International Publishing, 2021.

\bibitem{zavrtanik_draem_2021}
Vitjan Zavrtanik, Matej Kristan, and Danijel Skočaj.
\newblock {DRÆM} – a discriminatively trained reconstruction embedding for surface anomaly detection.
\newblock In {\em 2021 {IEEE}/{CVF} International Conference on Computer Vision ({ICCV})}, pages 8310--8319, 2021.
\newblock {ISSN}: 2380-7504.

\bibitem{zhou_extract_2022}
Chong Zhou, Chen~Change Loy, and Bo Dai.
\newblock Extract free dense labels from {CLIP}.
\newblock In Shai Avidan, Gabriel Brostow, Moustapha Cissé, Giovanni~Maria Farinella, and Tal Hassner, editors, {\em Computer Vision – {ECCV} 2022}, pages 696--712. Springer Nature Switzerland, 2022.

\bibitem{zhou_conditional_2022}
Kaiyang Zhou, Jingkang Yang, Chen~Change Loy, and Ziwei Liu.
\newblock Conditional prompt learning for vision-language models.
\newblock In {\em 2022 {IEEE}/{CVF} Conference on Computer Vision and Pattern Recognition ({CVPR})}, pages 16795--16804, 2022.
\newblock {ISSN}: 2575-7075.

\bibitem{zhou_anomalyclip_2023}
Qihang Zhou, Guansong Pang, Yu Tian, Shibo He, and Jiming Chen.
\newblock {AnomalyCLIP}: Object-agnostic prompt learning for zero-shot anomaly detection.
\newblock 2023.

\bibitem{zhu_toward_2024}
Jiawen Zhu and Guansong Pang.
\newblock Toward generalist anomaly detection via in-context residual learning with few-shot sample prompts, 2024.

\bibitem{zou_spot--difference_2022}
Yang Zou, Jongheon Jeong, Latha Pemula, Dongqing Zhang, and Onkar Dabeer.
\newblock {SPot}-the-difference self-supervised pre-training for anomaly detection and segmentation.
\newblock In Shai Avidan, Gabriel Brostow, Moustapha Cissé, Giovanni~Maria Farinella, and Tal Hassner, editors, {\em Computer Vision – {ECCV} 2022}, pages 392--408. Springer Nature Switzerland, 2022.

\end{thebibliography}
}

\clearpage
\appendix

\section{Details in Few-Shot Inference }
\label{sec:detail_fews_hot}

The HFWA inference framework, as illustrated in the Figure \ref{fig:figure 4}, integrates a Hierarchical Feature Memory Bank to store multi-level features from the normal reference images. The model processes the input image through the HFWA module, which captures hierarchical features at different levels. These features are then compared with the stored normal features to identify anomalies.

The comparison is based on the cosine similarity. Next, we aggregate all the anomaly maps to obtain the few-shot result $F_{few}$. Mathematically, this is represented as:
\begin{equation}
F_few = \sum^{4}_{l} \min_{f \in D_n} \left(1 - \langle F^{\text ref}_l, F_{\text fusion} \rangle \right).
\end{equation}
Here, $<\cdot, \cdot>$ represents the cosine similarity. Finally, the anomaly map $S_{few}$ obtained by feature comparison is combined with the anomaly map $F_{fusion}$ derived from zero-shot inference to produce the final result.

By incorporating these multi-level reference features, our extended HFWA model can more precisely define and recognize anomalies, leveraging the complementary predictions from both language-guided and visual-based approaches.


\section{Anomaly Detection Benchmark}
\label{sec:benchmark}

\begin{table}[t]
\centering
\caption{Datasets and their characteristics}
\label{table:table4}
\small 
\resizebox{\columnwidth}{!}{%
\begin{tabular}{ccccc}
\toprule
\textbf{Dataset} & \textbf{Category} & \textbf{Modalities} & \textbf{|C|} & \textbf{Normal/Anomalous} \\
\midrule
    MVTec AD & Obj \& texture & Photography & 15 & (467, 1258) \\
    VisA & Photography & Photography & 12 & (962, 1200) \\
    BTAD & Obj & Photography & 3 & (451, 290) \\
    DAGM & Texture & Photography & 10 & (6996, 1054) \\
    DTD-Synthetic & Photography & Photography & 12 & (357, 947) \\
\bottomrule
\end{tabular}
}
\end{table}

\noindent\textbf{MVTec-AD:} MVTec-AD contains 15 objects, with images having a resolution ranging from 7002 to 9002 pixels. Anomaly detection in this dataset is a one-class task, where the training set comprises only normal samples, while the test set includes both normal and anomalous samples. These samples are annotated with image-level and pixel-level labels. Additionally, the anomaly categories for each object are specified. This comprehensive annotation aids in developing and evaluating anomaly detection algorithms by providing clear distinctions between normal and abnormal conditions.

\noindent\textbf{VisA:} The VisA dataset includes 12 objects, each with images approximately 1.5K × 1K pixels in size. Similar to MVTec-AD, anomaly detection in VisA is a one-class task, with the training set containing only normal samples and the test set containing both normal and anomalous samples. These samples are annotated with image-level and pixel-level labels, along with the specific anomaly categories present for each object. This dataset is crucial for testing the robustness of anomaly detection models across different object types and anomaly scenarios.

\noindent\textbf{BTAD:} The BTAD (beanTech Anomaly Detection dataset) is a real-world industrial dataset designed for the anomaly detection task. It contains a total of 2830 real-world images of three industrial products, showcasing body and surface defects. The dataset provides a realistic scenario for testing anomaly detection algorithms, as it includes various types of defects that commonly occur in industrial settings. The BTAD dataset is essential for evaluating the performance of models in practical, real-world applications where the diversity and complexity of anomalies can significantly challenge detection capabilities.

\noindent\textbf{DAGM:} The DAGM dataset consists of artificially generated data that closely simulates real-world problems. It is divided into ten datasets, with the first six (development datasets) used for algorithm development and the remaining four (competition datasets) for performance evaluation. Each dataset comprises 1000 images showing background textures without defects and 150 images with one labeled defect per background texture. The images in each dataset are very similar, but each dataset is generated using different texture and defect models. This controlled environment allows for the systematic evaluation of anomaly detection algorithms under varying conditions.

\noindent\textbf{DTD-Synthetic:} DTD-Synthetic is a newly created dataset designed to test zero-shot anomaly detection methods on diverse data. It is based on the Describable Texture Dataset (DTD), which contains 47 texture classes, each with 120 different texture images, totaling 5,640 images. For DTD-Synthetic, twelve images were selected and synthesized to include anomalies, with resolutions ranging from 300 × 300 to 640 × 640 pixels. The resulting images, which range from 180 × 180 to 384 × 384 pixels, are generated by cropping regions with random orientations and positions. This dataset simulates factory image acquisition conditions and provides a variety of textures for robust anomaly detection testing.


\section{State-of-the-art Methods}
\label{sec:sota_methods}

In this study, we evaluate various state-of-the-art anomaly detection methods under different training settings as competing approaches. These settings include:

\noindent\textbf{SPADE:} This method uses alignment between an anomalous image and a set of similar normal images based on a multi-resolution feature pyramid, providing precise anomaly segmentation.

\noindent\textbf{PaDiM:} This framework utilizes a pretrained CNN for patch embedding and multivariate Gaussian distributions to probabilistically represent the normal class, enabling concurrent detection and localization of anomalies. 

\noindent\textbf{PatchCore:} This method utilizes a maximally representative memory bank of nominal patch features, combining embeddings from ImageNet models with outlier detection to address the cold-start problem in industrial anomaly detection.

\noindent\textbf{WinCLIP:} This method enhances CLIP for anomaly classification and segmentation by employing a compositional ensemble of state words and prompt templates. It efficiently extracts and aggregates window, patch, and image-level features aligned with text, addressing the limitations of CLIP in zero-shot and few-shot settings.

\noindent\textbf{April-GAN:} This method enhances the CLIP model for zero-shot anomaly detection by adding extra linear layers to map image features to a joint embedding space for comparison with text features, generating anomaly maps. 

\noindent\textbf{PromptAD:} This method introduces a one-class prompt learning approach for few-shot anomaly detection by using semantic concatenation to create anomaly prompts from normal prompts and constructing negative samples. It also employs an explicit anomaly margin to control the margin between normal and anomaly prompt features through a hyper-parameter, addressing training challenges caused by the absence of anomaly images.

\noindent\textbf{InCTRL:} This method trains a Generalist Anomaly Detection (GAD) model using few-shot normal images as sample prompts to generalize across diverse datasets without further training on the target data. It introduces an in-context residual learning model, trained on an auxiliary dataset, to discriminate anomalies from normal samples by evaluating the residuals between query images and few-shot normal sample prompts.


\section{Details in Performance Analysis}
\label{sec:detail_performance}

\begin{figure}[t]
\centering
\includegraphics[width=\columnwidth]{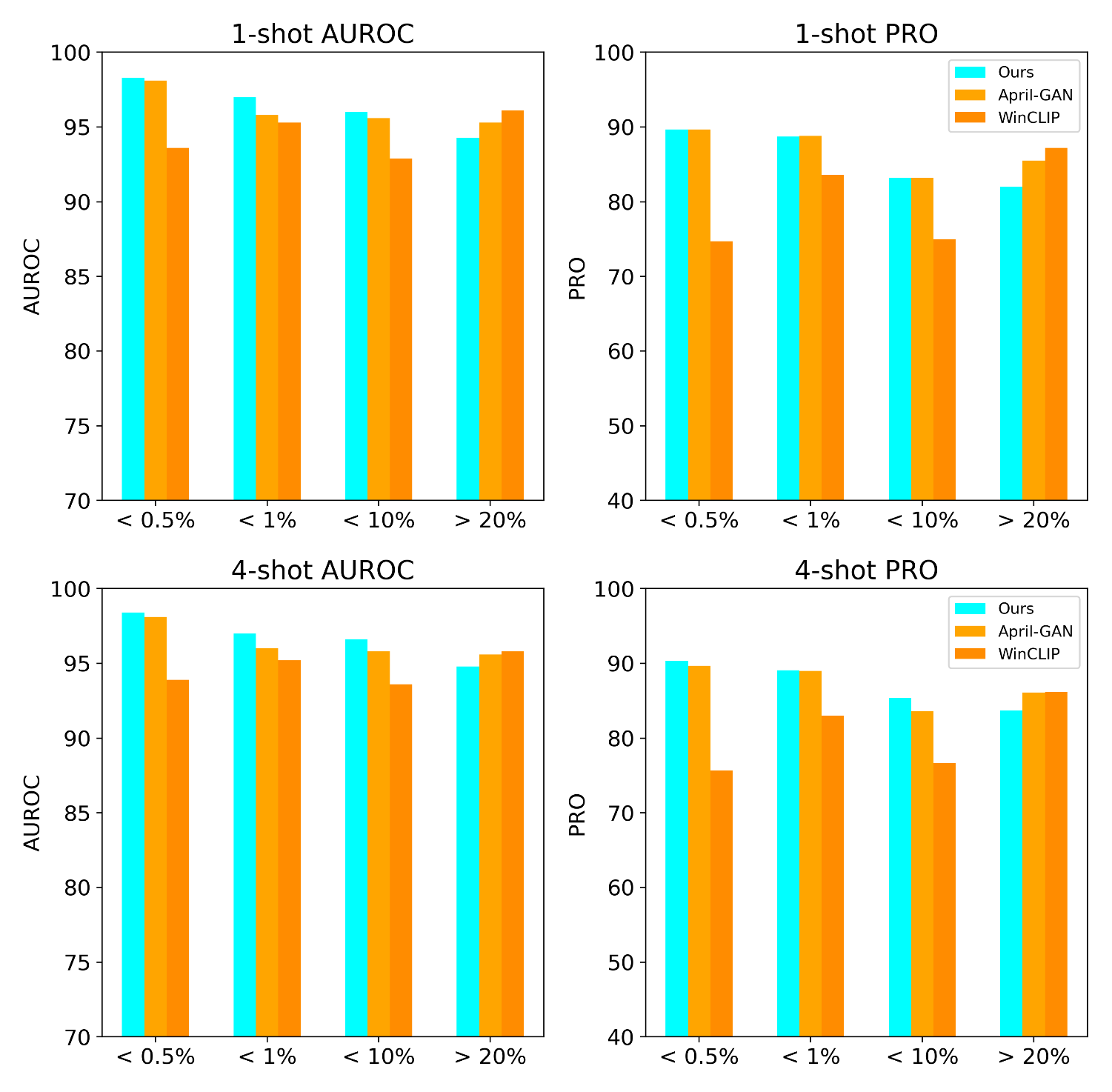}
\caption{Few-shot performance comparison with baselines.}
\label{fig:figure 10}
\end{figure}

Figure \ref{fig:figure 10} compares the few-shot metrics of our method with April-GAN and WinCLIP. Our method consistently outperforms April-GAN for anomalies less than 10\%, for anomalies greater than 10\%, April-GAN and WinCLIP perform better. 

\begin{figure*}[t]
\centering
\includegraphics[width=0.9\textwidth]{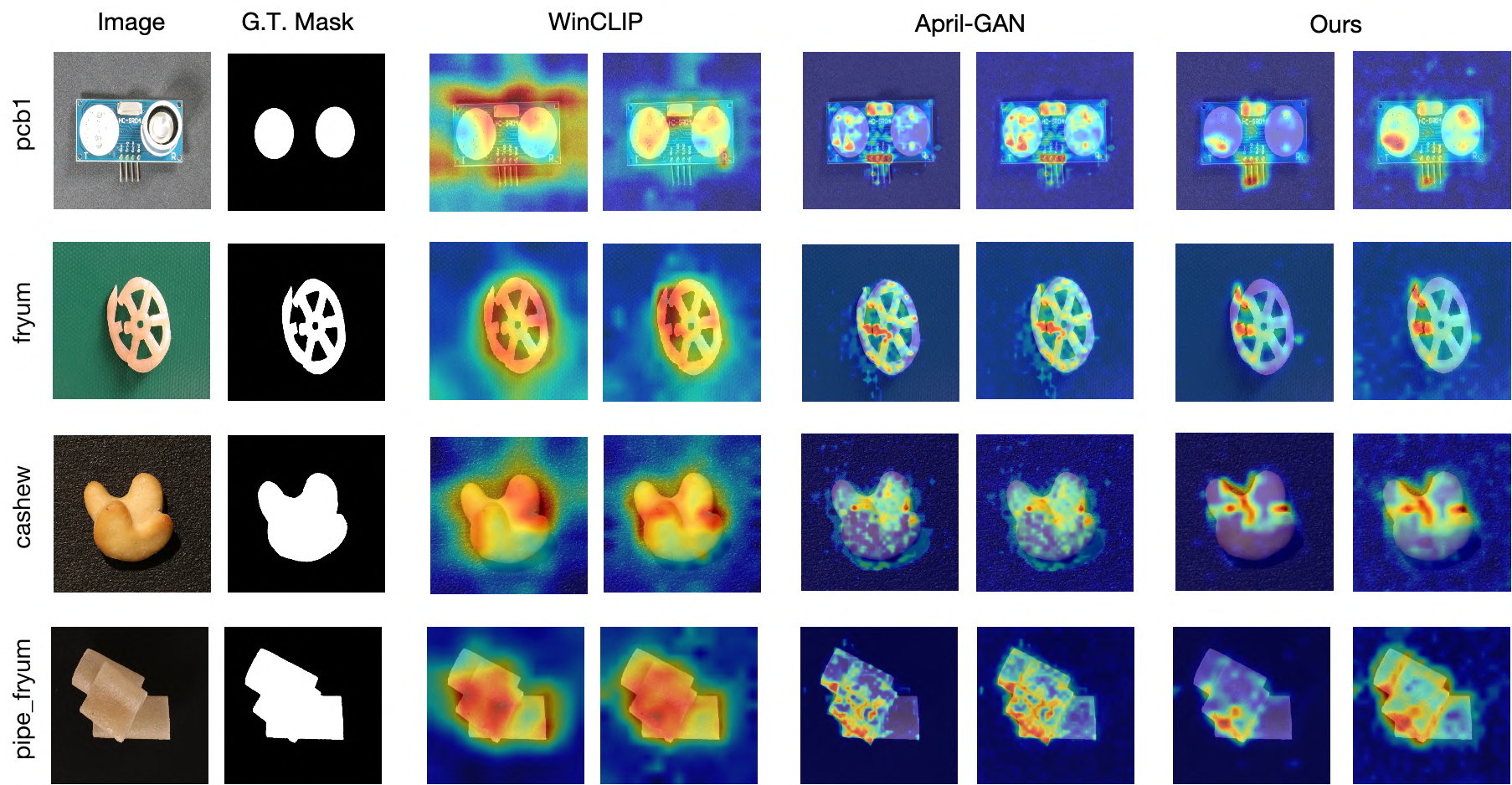}
\caption{Anomaly detection results on common and stacked normal objects when anomaly size exceeds 10\%.}
\label{fig:figure 11}
\end{figure*}

Figure \ref{fig:figure 11} shows visualization results when the anomaly size exceeds 10\%. For the top two rows, which represent common cases, our model accurately identifies and highlights the anomalous regions. The heatmaps correspond well to the actual anomalies in the original images, demonstrating the model's high precision in anomaly localization.

However, the bottom two rows present a more challenging scenario where normal objects are stacked together, creating multi-layered structures. In these cases, even human judgment would struggle to classify the situation as anomalous without additional rule-based semantic information. Typically, CLIP-based models, including ours, do not incorporate this extra semantic input.

Our model's failure in these scenarios is due to its working principle, which focuses on establishing connections between local and global anomaly features. Since the stacked objects are all normal, the model can only focus on normal edges and cannot establish a relationship based on anomaly features.

On the other hand, WinCLIP outperforms in these cases because it relies on global features from the last layer of ViT, which tends to overestimate the extent of anomalous regions. In zero-shot settings, WinCLIP significantly expands the estimation of anomaly features. When reference images are used, this overestimation is corrected, reducing the anomaly estimation area and improving performance.

In summary, due to the lack of additional rule-based semantic information, directly classifying stacked normal objects as anomalies is unreasonable. This highlights the need for incorporating additional semantic information and rule-based inputs to improve anomaly detection in such scenarios. While our model excels in standard anomaly detection, addressing the complexity of stacked objects requires a more sophisticated semantic understanding.


\section{Zero-shot Performance Comparison}
\label{sec:zero_shot}

\begin{table}[t]
\centering
\caption{Performance comparison across different datasets and methods under the Zero-shot setting. The best result for each dataset is highlighted in \textbf{\textcolor{red}{red}}.}
\label{table:table5}
\small 
\resizebox{\columnwidth}{!}{%
\begin{tabular}{l l c c c c}
\toprule
\textbf{Metric} & \textbf{Dataset} & \textbf{WinCLIP} & \textbf{April-GAN} & \textbf{AnormalyCLIP} & \textbf{Ours} \\
\midrule
\multirow{5}{*}{\textbf{AC AUROC}} 
    & MVTec-AD & \textbf{\textcolor{red}{91.8}} & 86.1 & 91.5 & 87.1 \\
    & Visa & 78.1 & 81.2 & \textbf{\textcolor{red}{82.1}} & 80.0 \\
    & BTAD & 68.2 & 73.6 & \textbf{\textcolor{red}{88.3}} & 76.4 \\
    & DAGM & 91.8 & 94.4 & \textbf{\textcolor{red}{97.5}} & 96.7 \\
    & DTD-Synthetic & 95.1 & 85.5 & 93.5 & \textbf{\textcolor{red}{95.2}} \\
\midrule
\multirow{5}{*}{\textbf{AC AP}} 
    & MVTec-AD & \textbf{\textcolor{red}{96.5}} & 93.5 & 96.2 & 93.7 \\
    & Visa & 78.0 & 81.4 & \textbf{\textcolor{red}{85.4}} & 84.6 \\
    & BTAD & 70.9 & 69.7 & \textbf{\textcolor{red}{87.3}} & 78.5 \\
    & DAGM & 79.5 & 83.9 & \textbf{\textcolor{red}{92.3}} & 89.4 \\
    & DTD-Synthetic & 97.7 & 94.0 & 97.0 & \textbf{\textcolor{red}{98.0}} \\
\midrule
\multirow{5}{*}{\textbf{AS AUROC}} 
    & MVTec-AD & 85.1 & 87.6 & \textbf{\textcolor{red}{91.1}} & 90.0 \\
    & Visa & 79.6 & 94.0 & \textbf{\textcolor{red}{95.5}} & 94.7 \\
    & BTAD & 72.7 & 89.3 & 94.2 & \textbf{\textcolor{red}{95.5}} \\
    & DAGM & 87.6 & 82.4 & 95.6 & \textbf{\textcolor{red}{97.4}} \\
    & DTD-Synthetic & 79.5 & 95.2 & 97.9 & \textbf{\textcolor{red}{98.9}} \\
\midrule
\multirow{5}{*}{\textbf{AS AUPRO}} 
    & MVTec-AD & 64.6 & 44.0 & 81.4 & \textbf{\textcolor{red}{84.0}} \\
    & Visa & 56.8 & 86.8 & 87.0 & \textbf{\textcolor{red}{88.4}} \\
    & BTAD & 27.3 & 68.8 & 74.8 & \textbf{\textcolor{red}{77.8}} \\
    & DAGM & 65.7 & 66.2 & 91.0 & \textbf{\textcolor{red}{93.8}} \\
    & DTD-Synthetic & 51.5 & 87.4 & 92.3 & \textbf{\textcolor{red}{96.1}} \\
\bottomrule
\end{tabular}
}
\end{table}

\begin{table}[t]
\centering
\caption{Ablation studies across different model configurations on the MVTec-AD dataset under 4-shot setting. The best result for each metric is highlighted in \textbf{\textcolor{red}{red}}.}
\label{table:table6}
\small 
\resizebox{\columnwidth}{!}{%
\begin{tabular}{l c c c c}
\toprule
\textbf{Model} & \textbf{AC AUROC} & \textbf{AC AP} & \textbf{AS AUROC} & \textbf{AS PRO} \\
\midrule
HFWA w/ Coop & \textbf{\textcolor{red}{96.8}} & \textbf{\textcolor{red}{98.3}} & \textbf{\textcolor{red}{95.7}} & \textbf{\textcolor{red}{92.4}} \\
HFWA w/ Template & 96.0 & 97.8 & 95.4 & 91.9 \\
HFWA w/ Prompt & 93.3 & 96.8 & 94.9 & 90.7 \\
Linear w/ Coop & 94.1 & 97.0 & 94.7 & 91.2 \\
Linear w/ Prompt & 93.6 & 96.9 & 95.1 & 91.0 \\
\bottomrule
\end{tabular}
}
\end{table}

\begin{table}[t]
\centering
\caption{Ablation studies across different model configurations on the Visa dataset under the 4-shot setting. The best result for each metric is highlighted in \textbf{\textcolor{red}{red}}.}
\label{table:table7}
\small 
\resizebox{\columnwidth}{!}{%
\begin{tabular}{l c c c c}
\toprule
\textbf{Model} & \textbf{AC AUROC} & \textbf{AC AP} & \textbf{AS AUROC} & \textbf{AS PRO} \\
\midrule
HFWA w/ Coop & \textbf{\textcolor{red}{92.9}} & \textbf{\textcolor{red}{94.5}} & \textbf{\textcolor{red}{97.1}} & \textbf{\textcolor{red}{91.4}} \\
HFWA w/ Template & 91.9 & 93.9 & 97.0 & 90.6 \\
HFWA w/ Prompt & 92.6 & \textbf{\textcolor{red}{94.5}} & \textbf{\textcolor{red}{97.1}} & 90.4 \\
Linear w/ Coop & 90.5 & 92.9 & 96.7 & 89.7 \\
Linear w/ Prompt & 92.5 & 94.4 & 97.0 & 90.6 \\
\bottomrule
\end{tabular}
}
\end{table}

\begin{figure*}[t] 
  \centering
  \begin{minipage}[t]{0.45\textwidth} 
    \centering
    \includegraphics[width=\textwidth]{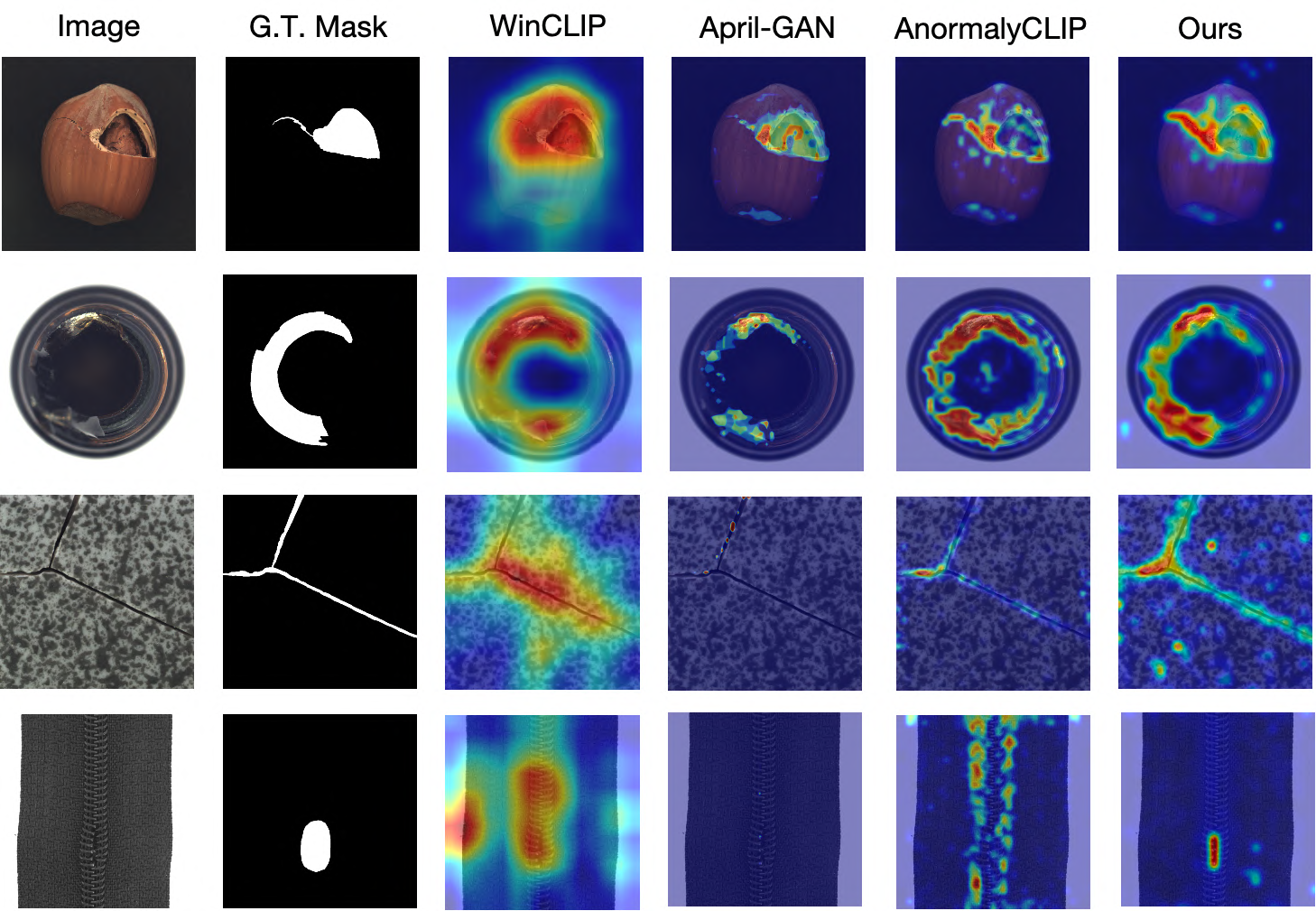}
    \caption{Visualization results on the MVTec-AD dataset under the Zero-shot setting.}
    \label{fig:figure 12}
  \end{minipage}%
  \hspace{0.05\textwidth} 
  \begin{minipage}[t]{0.45\textwidth}
    \centering
    \includegraphics[width=\textwidth]{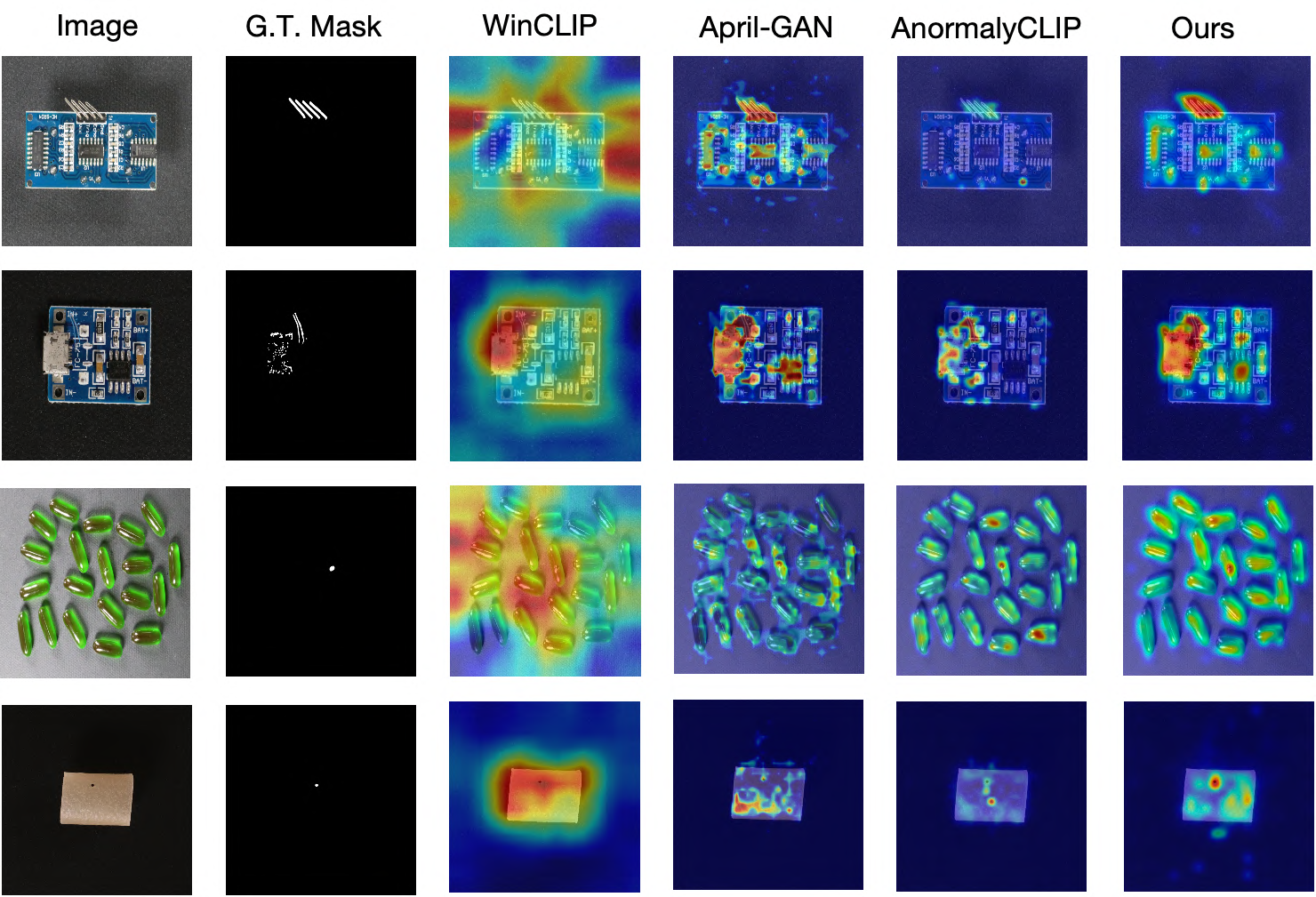}
    \caption{Visualization results on the VisA dataset under the Zero-shot setting.}
    \label{fig:figure 13}
  \end{minipage}
\end{figure*}

In the zero-shot setting, our analysis focuses on the performance of various models across different datasets for four key metrics: AC AUROC, AC AP, AS AUROC, and AS AUPRO. The models compared are WinCLIP, April-GAN, AnormalyCLIP \cite{zhou_anomalyclip_2023}, and our proposed model. The results are shown in Table \ref{fig:figure 5}. For AC AUROC, AnormalyCLIP consistently outperforms the other models on most datasets, notably achieving the highest scores on BTAD (88.3), DAGM (97.5), and Visa (82.1). Our model demonstrates competitive performance on DAGM (96.7) and DTD-Synthetic (95.2), but lags behind AnormalyCLIP on several datasets.

In terms of AC AP, AnormalyCLIP also excels, particularly on the BTAD (87.3), DAGM (92.3), and Visa (85.4) datasets. Our model shows strong results on DTD-Synthetic (98.0), surpassing AnormalyCLIP (97.0), and performing closely on DAGM (89.4). For AS AUROC, our model achieves the highest scores on BTAD (95.5) and DTD-Synthetic (98.9), indicating superior segmentation performance on these datasets. AnormalyCLIP performs best on MVTec-AD (91.1) and DAGM (95.6), while our model also shows competitive results on Visa (94.7).

Regarding AS AUPRO, our model outperforms others on DTD-Synthetic (96.1) and DAGM (93.8), highlighting its effectiveness in fine-grained anomaly segmentation. AnormalyCLIP performs well on MVTec-AD (81.4) and Visa (87.0), but our model achieves the highest score on Visa (88.4), demonstrating robustness in this setting. Overall, AnormalyCLIP generally shows strong performance in the zero-shot setting, particularly in AC AUROC and AC AP metrics. Our proposed model, however, excels in AS AUROC and AS AUPRO, especially on BTAD and DTD-Synthetic datasets, indicating its strength in anomaly segmentation tasks. These results underscore the importance of considering both classification and segmentation metrics when evaluating anomaly detection models in zero-shot scenarios.


\section{Ablation Studies}
\label{sec:ablation}

In our ablation studies, we investigate the performance of various model configurations on the MVTec-AD and Visa datasets. Our proposed model structure, HFWA w/ Coop, utilizes DualCoOp mechanisms, while HFWA w/ Template enhances this with more refined prompt templates. HFWA w/ Prompt and Linear models represent configurations without DualCoOp mechanisms, akin to WinCLIP and April-GAN, using non-trainable text encodings. The Linear configurations replace HFWA with linear layers.

The performance comparison on the MVTec-AD dataset (Table \ref{table:table5}) shows that HFWA w/ Coop consistently achieves the highest scores across all metrics, indicating its superior capability in anomaly detection. Specifically, it achieves an AC AUROC of 96.8, AC AP of 98.3, AS AUROC of 95.7, and AS PRO of 92.4.
On the Visa dataset (Table \ref{table:table6}), HFWA w/ Coop again demonstrates robust performance, achieving the highest AS AUROC (97.1) and AS PRO (91.4). HFWA w/ Prompt matches the top AC AP (94.5) and AS AUROC (97.1) but slightly lags in AS PRO (90.4), indicating that while prompt-based configurations maintain high detection accuracy, they may have room for improvement in precision.

Overall, the HFWA w/ Coop model stands out as the most effective configuration for anomaly detection on both datasets, underscoring the importance of hierarchical feature adaptation and learnable prompts in enhancing model performance for complex anomaly detection tasks.


\end{document}